%% file: main.tex
\input{preamble}

\input{publication-links}

\begin{document}

\bstctlcite{IEEEtran:BSTcontrol}

\title{Auditing the Judge:\\
Internal Signals for LLM Reasoning Assurance}

\author{\IEEEauthorblockN{Mayank Gulati and Gerhard Wunder}
  \IEEEauthorblockA{Freie Universit\"at Berlin\\
  \{mayank.gulati, gerhard.wunder\}@fu-berlin.de}\thanks{Accepted for publication at IEEE TPS-ISA 2026.
\copyright~2026 IEEE. Personal use of this material is permitted.
Permission from IEEE must be obtained for all other uses, in any current or
future media, including reprinting/republishing this material for advertising
or promotional purposes, creating new collective works, for resale or
redistribution to servers or lists, or reuse of any copyrighted component of
this work in other works.}}

\maketitle

\input{sections/00-abstract}


\input{sections/01-introduction}
\input{sections/02-related-work}
\input{sections/03-setting}
\input{sections/04-method}
\input{sections/05-setup}
\input{sections/06-results}
\input{sections/07-trust-boundary}
\input{sections/08-limitations}
\input{sections/09-future-work}
\input{sections/09-conclusion}

\section*{Appendices and Artifact}
Detailed proofs, supplementary analyses, and reproducibility documentation
are included in the appendices of this complete author version.
The software, cached results, and aligned predictions are available in the
\href{\releaseurl}{versioned public artifact release}.

\section*{Acknowledgment}
Mayank Gulati and Gerhard Wunder were supported by the Federal Ministry of Education and Research of
Germany (BMBF) in the programme of ``Souver\"an. Digital. Vernetzt.'', joint
project ``AIgenCY: Chances und Risks of Generative AI in Cybersecurity'', project
identification number 16KIS2013, and by BMBF joint project ``6G-RIC: 6G Research
and Innovation Cluster'', project identification number 16KISK025 16KISK020K,
as well as the BMBF joint project ``UltraSec: Security Architecture for UWB-based
Application Platform'', project identification number 16KIS1682. Gerhard Wunder was also
supported by the German Science Foundation (DFG) within priority program
SPP~2378: ``ResNets: Resilience in Connected Worlds'' under grant WU~598/12-1.

\emph{AI Disclosure: Large language model tools were used for language editing
and drafting assistance. All ideas, experiments, and results are our own; the
authors verified the correctness and originality of all content and take full
responsibility for the paper.}

\bibliographystyle{IEEEtran}
\bibliography{ref}

\clearpage
\appendices
\input{appendix/A-proofs}
\input{appendix/B-protocol}
\input{appendix/C-pyes-audit}
\input{appendix/D-populations}
\input{appendix/E-ablations}
\input{appendix/G-reproducibility}

\end{document}

%% file: preamble.tex
\documentclass[conference]{IEEEtran}
\IEEEoverridecommandlockouts

\usepackage{cite}
\usepackage{amsmath,amssymb,amsfonts,amsthm}
\usepackage{algorithmic}
\usepackage{graphicx}
\usepackage{textcomp}
\usepackage{xcolor}
\usepackage{booktabs}
\usepackage{multirow}
\usepackage{siunitx}
\usepackage{tikz}
\usetikzlibrary{arrows.meta, positioning, calc, fit, backgrounds}
\usepackage[hyphens]{url}
\usepackage{xr-hyper}
\usepackage[colorlinks=true,allcolors={blue!65!black},breaklinks=true]{hyperref}
\hypersetup{pdfauthor={Mayank Gulati and Gerhard Wunder}}
\usepackage[capitalise,noabbrev]{cleveref}

\graphicspath{{figures/}}

\input{macros}

%% file: macros.tex
\newtheorem{proposition}{Proposition}
\newtheorem{lemma}{Lemma}

\theoremstyle{definition}

\theoremstyle{remark}

\newcommand{\Prob}{\mathbb{P}}
\newcommand{\E}{\mathbb{E}}

\newcommand{\Real}{\mathbb{R}}
\newcommand{\indic}[1]{\mathbf{1}\!\left[#1\right]}
\newcommand{\Dtr}{\mathcal{D}_{\mathrm{tr}}}
\newcommand{\Dte}{\mathcal{D}_{\mathrm{te}}}

\newcommand{\query}{x}
\newcommand{\sol}{z}
\newcommand{\gold}{a^{\star}}
\newcommand{\pred}{\hat{a}}
\newcommand{\score}{s}
\newcommand{\prev}{\pi_{0}}          
\newcommand{\prevplus}{\pi_{+}}       
\newcommand{\prevminus}{\pi_{-}}      

\newcommand{\budget}{m}
\newcommand{\flagset}[1]{\mathcal{F}_{#1}}
\newcommand{\recall}{\mathcal{R}}     
\newcommand{\resid}{\varrho}          
\newcommand{\srisk}{\mathcal{S}}      
\newcommand{\auroc}{\mathrm{AUROC}}
\newcommand{\aurc}{\mathrm{AURC}}

\newcommand{\repair}{\gamma}          
\newcommand{\routeset}{\mathcal{A}}
\newcommand{\acc}{\mathrm{Acc}}

\newcommand{\acts}{A}
\newcommand{\feats}{F}
\newcommand{\resfeats}{R}
\newcommand{\heads}{\mathcal{T}}
\newcommand{\headscore}{\alpha}
\newcommand{\Bhat}{\widehat{B}}
\newcommand{\phihat}{\widehat{\phi}}
\newcommand{\pyes}{p(\textsc{yes})}
\newcommand{\yes}{\textsc{yes}}
\newcommand{\no}{\textsc{no}}

\newcommand{\srcdom}{\mathrm{S}}
\newcommand{\tgtdom}{\mathrm{T}}

\definecolor{oiblue}{RGB}{0,114,178}
\definecolor{oiorange}{RGB}{230,159,0}
\definecolor{oigreen}{RGB}{0,158,115}
\definecolor{oivermilion}{RGB}{213,94,0}
\definecolor{oipurple}{RGB}{204,121,167}

%% file: publication-links.tex
\newcommand{\releaseurl}{https://github.com/MakGulati/auditing-the-judge-artifact/releases/tag/tps2026-v1}
\newcommand{\supplementurl}{https://github.com/MakGulati/auditing-the-judge-artifact/releases/download/tps2026-v1/supplement.pdf}

%% file: sections/00-abstract.tex
\begin{abstract}
Language-model systems use self-judgment to decide which answers to release,
review, or recompute. Because intervention is costly, an assurance score must
rank errors well at the available budget, not merely over the full ranking. We
compare the judge's continuous verdict-token score with a supervised correctness
probe trained on attention-head activations from the same prompt. Exact
identities link fixed-budget error recall to residual risk and show that AUROC
averages recall over all budgets. Across three open 7--12B models on GSM8K and
MATH, the probe finds \num{1.3}--\num{22.5} additional percentage points of
errors at 5\% review in all six model--dataset settings (median \num{7.4}
points). At 10\% routing,
probe-guided allocation improves accuracy over equal-cost random allocation in
both GSM8K experiments, but beats judge-guided routing only for the weaker judge.
Cross-dataset transfer lowers probe AUROC for every model in both directions
relative to target-domain training. Because dataset, weight precision, and
generation budget change together, this measures a joint protocol shift rather
than a dataset-only effect. Internal signals can improve supervised, in-domain
budget allocation, but are not domain-general confidence estimators.
\end{abstract}

%% file: sections/01-introduction.tex
\section{Introduction}
\label{sec:intro}

Self-evaluation becomes an assurance mechanism when a model's judgment determines
which answers receive further scrutiny. A deployer may send low-scoring answers
to a human or rerun them with additional inference, while releasing the
rest~\cite{zheng2023judging,bai2022constitutional,yuan2024selfrewarding}. An
incorrect answer assigned a high score can therefore bypass intervention and
reach the user. Because human attention and inference compute are limited, the
deployment cannot scrutinize every answer. It must spend its budget on the
outputs most likely to be wrong. The central question is thus which
self-evaluation signal best allocates a fixed intervention budget.

We position the probe as a deployment audit tool, not a correctness guarantee:
an auditor fixes a review budget, validates the score on labelled outputs from
the intended protocol, and uses its ranking to prioritise human review or
recomputation while measuring residual error among released answers. This
operational use is consistent with risk-management practices that emphasise
measurement, monitoring, documentation, and human oversight, including the NIST
AI Risk Management Framework and, where applicable, the EU AI
Act~\cite{nist2023airmf,eu2024aiact}. The score can support such controls, but it
does not establish regulatory compliance and must be revalidated after a change
to the model, data, or inference protocol.

A verbalized judgment may discard information useful for that allocation. The
judging pass builds a high-dimensional internal state, then compresses it into a
verdict token. Even the continuous \yes/\no\ token mass is only one projection of
that state. Prior work shows that internal activations can contain
correctness-related information~\cite{azaria2023internal,marks2024geometry,
li2024inference}. However, aggregate discrimination does not establish
deployment value. It does not show how many errors a small review budget catches,
how much accuracy a compute budget buys, or whether a fitted score remains valid
after the deployment protocol changes.

Standard ranking metrics can obscure this distinction. AUROC summarizes the
entire ordering, while a deployment acts at one operating budget.
\Cref{prop:auroc} makes this mismatch precise: at fixed error prevalence, AUROC
is an affine function of error recall averaged uniformly over all review
budgets. It therefore weighs an 80\% review budget as heavily as a 5\% budget.
On Qwen2.5/GSM8K, the paired AUROC comparison does not establish a difference
between the probe and judge. Yet at 5\% review, the probe catches $9.3$
additional percentage points of the errors
(\cref{tab:main} and \cref{tab:budget}). This is the decision that matters when
only one answer in twenty can be reviewed.

Our core idea is to compare two read-outs of the same judging computation. The
black-box score is the judge's continuous $\pyes$ token mass. The white-box score
is a supervised probe fitted to attention-head activations at the verdict
position. Head selection and probe fitting use only training data. On held-out
answers, both scores rank the same population and receive the same review or
compute budget. This isolates whether the internal state improves the deployment
decision.

Our contributions are:

\begin{itemize}
  \item \textbf{Formalize budgeted assurance.} We show that human review and
        additional inference are fixed-budget selection problems. Exact results
        connect the selected ranking to error recall, residual risk, and routing
        accuracy.
  \item \textbf{Measure deployment value.} The probe improves error capture at
        5\% review in all six in-domain settings. It also beats equal-cost random
        routing in both compute experiments, but beats the judge only when the
        judge is weaker.
  \item \textbf{Identify the validity boundary.} Cross-dataset transfer reduces
        probe AUROC in every evaluated direction. Because these comparisons also
        change weight precision and generation budget, they reveal sensitivity
        to a joint protocol shift and motivate target-protocol validation.
\end{itemize}

The strongest supported claim is conditional. With supervision and validation
on the target model, task, and inference protocol, internal signals can improve
how limited review and compute are allocated. They do not provide domain-general
confidence or robustness to an active adversary.

%% file: sections/02-related-work.tex
\section{Related Work}
\label{sec:related}

\paragraph*{Model-based evaluation and self-correction}
LLMs serve as evaluators in comparative benchmarks~\cite{zheng2023judging}, as
sources of feedback during training~\cite{bai2022constitutional,
yuan2024selfrewarding}, and as critics in iterative refinement
~\cite{madaan2023selfrefine}. These uses depend on the model recognizing defects
in generated text. That ability is imperfect: model judges exhibit position,
verbosity, and self-preference biases~\cite{zheng2023judging,
panickssery2024llm,wang2024fairness}, while self-correction without external
feedback often fails to improve reasoning~\cite{huang2024largecannot,
kamoi2024cansurvey}. We therefore compare the probe with a continuous black-box
self-judgment score: the judge's normalized \yes/\no\ token probability, rather
than a sampled binary verdict. This is a strong baseline from the same judging
pass, although it does not exhaust output-only confidence methods.

\paragraph*{Selective prediction and abstention}
Selective prediction formalizes the option to reject an uncertain output
~\cite{chow1970optimum,elyaniv2010foundations}. Later work develops risk--coverage
metrics and learned rejectors for deep networks~\cite{geifman2017selective,
geifman2019selectivenet,zhou2025aurc}, as well as misclassification detection
~\cite{hendrycks2017baseline}. For language models, selective generation handles
distribution shift~\cite{ren2023outofdistribution}, and selective answering
handles ambiguous questions~\cite{cole2023selectively}. We use the same
operational view: a score ranks inputs for review under a fixed budget. This
budget matters because AUROC averages error-detection performance over all
possible review fractions (\cref{prop:auroc}), whereas a deployment operates at
one or a few chosen fractions.

\paragraph*{Internal signals of truth and correctness}
Linear and shallow probes recover truth-related structure from transformer hidden
states~\cite{azaria2023internal,burns2022discovering,marks2024geometry}.
Internal states have also supported confidence estimation across layers
~\cite{beigi2024internalinspector}, long-form factuality detection
~\cite{han2025simple}, and multi-sample hallucination detection
~\cite{chen2024insideout}; attention-head interventions can additionally steer
truthful answers~\cite{li2024inference}. Output-side alternatives instead compare
the meanings or agreement of sampled generations~\cite{kuhn2023semantic,
farquhar2024detecting,manakul2023selfcheckgpt}. Our probe differs by reading the
hidden state of a separate \emph{judging} pass and by evaluating fixed review
budgets and cross-dataset transfer. This design creates an interpretive
ambiguity. Objective correctness and self-judgment usually agree, so
correctness-labelled training data cannot distinguish a probe that tracks actual
correctness from one that tracks the judge's belief~\cite{lu2026diagnosing}. Our
results therefore show that the probe predicts errors, but not which of these two
signals it has learned.

\paragraph*{Adaptive inference}
Self-consistency improves reasoning by sampling several solution paths and
selecting their consensus~\cite{wang2023selfconsistency}; more generally,
test-time compute can trade inference cost for accuracy~\cite{snell2025scaling}.
Adaptive methods stop sampling when answers agree~\cite{aggarwal2023adaptive} or
learn how much compute to allocate to each input~\cite{damani2025learning}. Our
routing experiment asks a narrower question: can a judge-derived score select
which inputs receive $k$-sample self-consistency? The equal-cost random-routing
baseline isolates the value of that selection score from the decoder's average
gain (\cref{eq:route-random}).

\paragraph*{Benchmark validity}
GSM8K and MATH~\cite{cobbe2021gsm8k,hendrycks2021math} predate the checkpoints
studied here, so benchmark exposure is a validity concern. GSM1k tests for
contamination-related overfitting with newly written problems that match GSM8K's
style and difficulty~\cite{zhang2024careful}. GSM-Symbolic instead varies names,
numbers, and clauses within symbolic templates to test reasoning robustness
~\cite{mirzadeh2025gsmsymbolic}. Our GSM-Symbolic-style perturbation serves only
as a sensitivity analysis; it cannot establish that the original items were
absent from pretraining (\cref{app:symbolic}).

%% file: sections/03-setting.tex
\section{Budgeted Assurance Setting}
\label{sec:setting}

The deployment pipeline is
\emph{solve $\rightarrow$ score $\rightarrow$ select $\rightarrow$ intervene
$\rightarrow$ release}. The model produces an answer and a self-evaluation
score; the deployer uses that score to decide which answers receive a scarce
intervention. The score therefore informs an allocation decision but does not
itself establish correctness. This section formalizes that decision, the two
interventions studied here, and the exact quantities used to evaluate them. The
formal results make each reported quantity auditable; their proofs and
artifact-level identity checks are separated in \cref{app:proofs}.

\subsection{Answers, labels, and scores}
\label{sec:setting-task}

For a query $\query$, the solver emits a solution $\sol=(r,\pred)$ with reasoning
trace $r$ and final answer $\pred$. Let $\gold(\query)$ be the reference answer
and let $\equiv$ denote the dataset-specific answer-equivalence relation
(\cref{sec:setup-data}). Correctness is
\begin{equation}
  \label{eq:label}
  y(\query,\sol) = \indic{\pred \equiv \gold(\query)} \in \{0,1\},
\end{equation}
so $y=0$ denotes an error.

All learned components are fitted on $\Dtr$. Operational quantities are computed
on the held-out set
$\Dte=\{(\query_i,\sol_i,y_i)\}_{i=1}^{N}$. Let
$I\sim\operatorname{Unif}([N])$, where $[N]=\{1,\ldots,N\}$, denote a uniformly
sampled held-out item. We use $\Prob_N$ and $\E_N$ for probability and expectation
over $I$ and the tie-breaking rule defined below. The observed error prevalence is
\begin{equation*}
  \prev = \Prob_N(y_I=0)
  = \frac{1}{N}\sum_{i=1}^{N}\indic{y_i=0}.
\end{equation*}
This empirical probability space is used for the finite identities below; the
population statement in \cref{prop:auroc} is identified explicitly.

The model does not emit an authoritative correctness label. Instead, the control
supplies a real-valued score
\begin{equation}
  \label{eq:score}
  \score:(\query,\sol)\mapsto\Real,
\end{equation}
where larger values indicate greater confidence in correctness. We compare a
black-box score obtained from the judge's verdict-token distribution with a
white-box score learned from its internal activations (\cref{sec:method}).

\subsection{Fixed-budget selection}
\label{sec:setting-levers}

The deployer has capacity to intervene on an integer $M_\budget\in\{0,\ldots,N\}$
of the $N$ answers. The nominal budget is $\budget\in[0,1]$, and
$b_\budget=M_\budget/N$ is the realized fraction after the experiment-specific
rounding rule (\cref{sec:setup-stats}). A score selects the $M_\budget$
lowest-scoring answers,
\begin{equation}
  \label{eq:flagset}
  \begin{aligned}
    &\flagset{\budget}(\score)\subseteq[N],
    &\big|\flagset{\budget}(\score)\big|&=M_\budget,\\
    &\score_i\le\score_{i'}
    &\forall i\in\flagset{\budget}(\score),\quad
      &i'\notin\flagset{\budget}(\score).
  \end{aligned}
\end{equation}
If several items have the same score at the cutoff, we choose uniformly at random
among them to fill the remaining slots. Reported metrics average over this random
choice (\cref{app:ties}).

Only the ordering induced by $\score$ enters \cref{eq:flagset}. Consequently,
every strictly increasing transformation of the score produces the same decision.
Calibration can matter for other policies, but it is not required for this
fixed-capacity ranking problem.

We study two interventions with a fixed cost per selected item:
\begin{enumerate}
  \item \textbf{Review.} The selected answers are reviewed before release; the
        capacity $M_\budget$ is measured in answers.
  \item \textbf{Additional inference.} The selected queries receive $k$ new
        sampled solutions and a majority-vote answer. Because $k$ is fixed,
        selecting $M_\budget$ queries consumes $kM_\budget$ additional solver
        calls.
\end{enumerate}
The common decision is which subset receives the intervention. Its value,
however, differs between review and recomputation.

\subsection{Review objective: catch errors before release}
\label{sec:setting-review}

For review, the relevant outcomes are the fraction of errors selected and the
error rate among answers released without review:
\begin{align}
  \label{eq:recall}
  \recall_{\budget}(\score)
  &= \Prob_N\!\left(I\in\flagset{\budget}(\score)\mid y_I=0\right), \\
  \label{eq:residual}
  \resid_{\budget}(\score)
  &= \Prob_N\!\left(y_I=0\mid I\notin\flagset{\budget}(\score)\right).
\end{align}
The first is error recall at the review budget; the second is residual risk among
unreviewed answers. At a fixed budget and prevalence, they are two views of the
same error count.

\begin{proposition}[Finite-sample residual-risk identity]
  \label{prop:residual}
  Suppose the held-out set contains at least one error and $M_\budget<N$. For any
  score $\score$,
  \begin{equation}
    \label{eq:residual-identity}
    \resid_{\budget}(\score)
    = \prev\,\frac{1-\recall_{\budget}(\score)}{1-b_\budget}.
  \end{equation}
  Hence, for scores $\score$ and $\score'$ evaluated on the same held-out set at
  the same realized budget,
  \begin{equation}
    \label{eq:residual-delta}
    \resid_{\budget}(\score')-\resid_{\budget}(\score)
    = \frac{\prev}{1-b_\budget}
      \left[\recall_{\budget}(\score)-\recall_{\budget}(\score')\right].
  \end{equation}
\end{proposition}

The proof is a counting argument (\cref{app:proof-residual}). The identity has
two reporting consequences. First, recall and residual risk are not independent
evidence: once $\prev$ and $b_\budget$ are fixed, either determines the other.
Second, the conversion from a recall gain to a residual-risk reduction scales as
$\prev/(1-b_\budget)$. Residual-risk gains should therefore be compared between
scores on the same population, not across populations with different error
prevalence. At full review, $b_\budget=1$ and no unreviewed answers remain, so
residual risk is undefined.

\subsection{Why aggregate ranking can miss the operating point}
\label{sec:setting-auroc}

AUROC summarizes the full ranking, whereas deployment uses a specific budget.
For a population budget $\budget$, select the lowest-scoring fraction $\budget$.
If items tie at the cutoff, select each with equal probability so that the total
selected fraction remains $\budget$. Let $\recall^{P}_{\budget}(\score)$ be the
resulting error recall.

\begin{proposition}[AUROC averages recall over budgets]
  \label{prop:auroc}
  Let $Y$ be the correctness label for an answer drawn from a population $P$
  containing both correct and incorrect answers, and let
  $\pi_{0,P}=\Prob_P(Y=0)$. Orient $\score$ so that larger values indicate greater
  confidence in correctness. To compute $\auroc_P(\score)$, compare an
  independently drawn correct and incorrect answer: count one if the correct
  answer scores higher, zero if it scores lower, and one half if the scores are
  equal. Then
  \begin{equation}
    \label{eq:auroc-identity}
    \int_{0}^{1}\recall^{P}_{\budget}(\score)\,\mathrm{d}\budget
    = \frac{\pi_{0,P}}{2}
      +(1-\pi_{0,P})\auroc_P(\score).
  \end{equation}
\end{proposition}

The proof is given in \cref{app:proof-auroc}. At a fixed error prevalence, the
identity relates AUROC to error recall averaged over all budgets from zero to one.
It therefore gives equal weight to equal-sized budget ranges, even when deployment
uses only a small budget. Two scores can consequently have similar AUROC but
different error recall at the deployment budget.

We report AUROC for comparability, but state deployment claims at fixed budgets.
The exact finite-sample analogue and the corresponding risk--coverage view are
given in \cref{app:proof-auroc} and \cref{app:risk-coverage}.

\subsection{Compute objective: allocate inference where it changes the answer}
\label{sec:setting-routing}

Additional inference is valuable only when it changes correctness. Condition on
the realized greedy and $k$-sample majority-vote outputs. For item $i$, define
the signed \emph{repair value}
\begin{equation}
  \label{eq:repair}
  \repair_i
  = \indic{\pred^{(k)}_i\equiv\gold(\query_i)}
    -\indic{\pred^{(1)}_i\equiv\gold(\query_i)}
  \in\{-1,0,+1\}.
\end{equation}
A value of $+1$ means that additional inference repairs an error, $-1$ means it
breaks a correct answer, and $0$ means correctness is unchanged. Routing a set
$\routeset\subseteq[N]$ to the expensive decoder gives
\begin{equation}
  \label{eq:route-acc}
  \acc(\routeset)
  = \acc_{\mathrm{greedy}}+\frac{1}{N}\sum_{i\in\routeset}\repair_i.
\end{equation}
Its implemented cost is $1+k|\routeset|/N$ solver calls per query: one initial
greedy solve for every query and $k$ new solves for every routed query. The judge
pass is outside this solve-call metric.

We first compare score-guided routing with random routing at the same budget. Let
$\bar\repair=N^{-1}\sum_i\repair_i$ be the average change in correctness when
additional inference is applied across all queries. If $M_\budget$ queries are
selected uniformly at random, the expected accuracy is
\begin{equation}
  \label{eq:route-random}
  \acc_{\mathrm{rand}}(\budget)
  :=\E_{\routeset}[\acc(\routeset)]
  =\acc_{\mathrm{greedy}}+b_\budget\bar\repair.
\end{equation}
If $\bar\repair>0$, random routing already improves on greedy decoding. This gain
comes from additional inference, not from the score. The score adds selection
value only if it beats random routing at the same cost.

An oracle provides an upper bound. It observes each $\repair_i$ and then selects
the best $M_\budget$ queries. Let
$\prevplus=\Prob_N(\repair_I=+1)$ and
$\prevminus=\Prob_N(\repair_I=-1)$ be the fractions of queries helped and harmed
by additional inference. Its accuracy is
\begin{align}
  \label{eq:route-oracle}
  \acc_{\mathrm{orac}}(\budget)
  &:=\max_{|\routeset|=M_\budget}\acc(\routeset) \notag\\
  &=\acc_{\mathrm{greedy}}+\min(b_\budget,\prevplus) \notag\\
  &\quad-\max\!\left\{0,b_\budget-(1-\prevminus)\right\}.
\end{align}
This expression assumes that exactly $M_\budget$ queries are routed, as in our
experiments. If $M_\budget$ is only an upper limit, the oracle never routes harmed
queries and the final subtraction term disappears.
The oracle routes helped queries first and harmed queries only when needed to fill
the budget. Its decisions require outcomes observed after additional decoding, so
it cannot be deployed. Both identities are proved in \cref{app:proof-routing}.

A deployed policy must choose before the additional samples are drawn. Let
$\mathcal I_i$ denote the information available at that point. Assuming that
routing one query does not affect another, the best fixed-budget policy ranks
queries by their expected repair value,
$\E[\repair_i\mid\mathcal I_i]$. A correctness score instead estimates whether
the current answer is correct. These targets differ because an incorrect answer
may remain incorrect after additional inference.

\subsection{Assurance criteria and scope}
\label{sec:setting-trust}

A score is useful only if it improves the budgeted decision. For review, we
compare scores on the same held-out population and at the same budget. A better
score has higher $\recall_\budget$, equivalently lower $\resid_\budget$. For
compute, score-guided routing must first outperform equal-cost random routing;
comparison with judge-guided routing then tests whether internal access adds value
beyond the black-box self-judgment score.

These are empirical, protocol-specific claims rather than guarantees. They apply
to the evaluated model, task, generation and judging protocol, label rule, and
budget, with the probe fitted on labeled in-domain answers. We assume a
non-adaptive query population and fixed per-item intervention costs. We do not
consider an adversary who manipulates queries, solutions, scores, or the review
process. Cross-protocol behavior and the cost of in-domain labels are examined in
\cref{sec:trust-boundary} and \cref{sec:limitations}.

%% file: sections/04-method.tex
\section{Constructing the Correctness Score}
\label{sec:method}

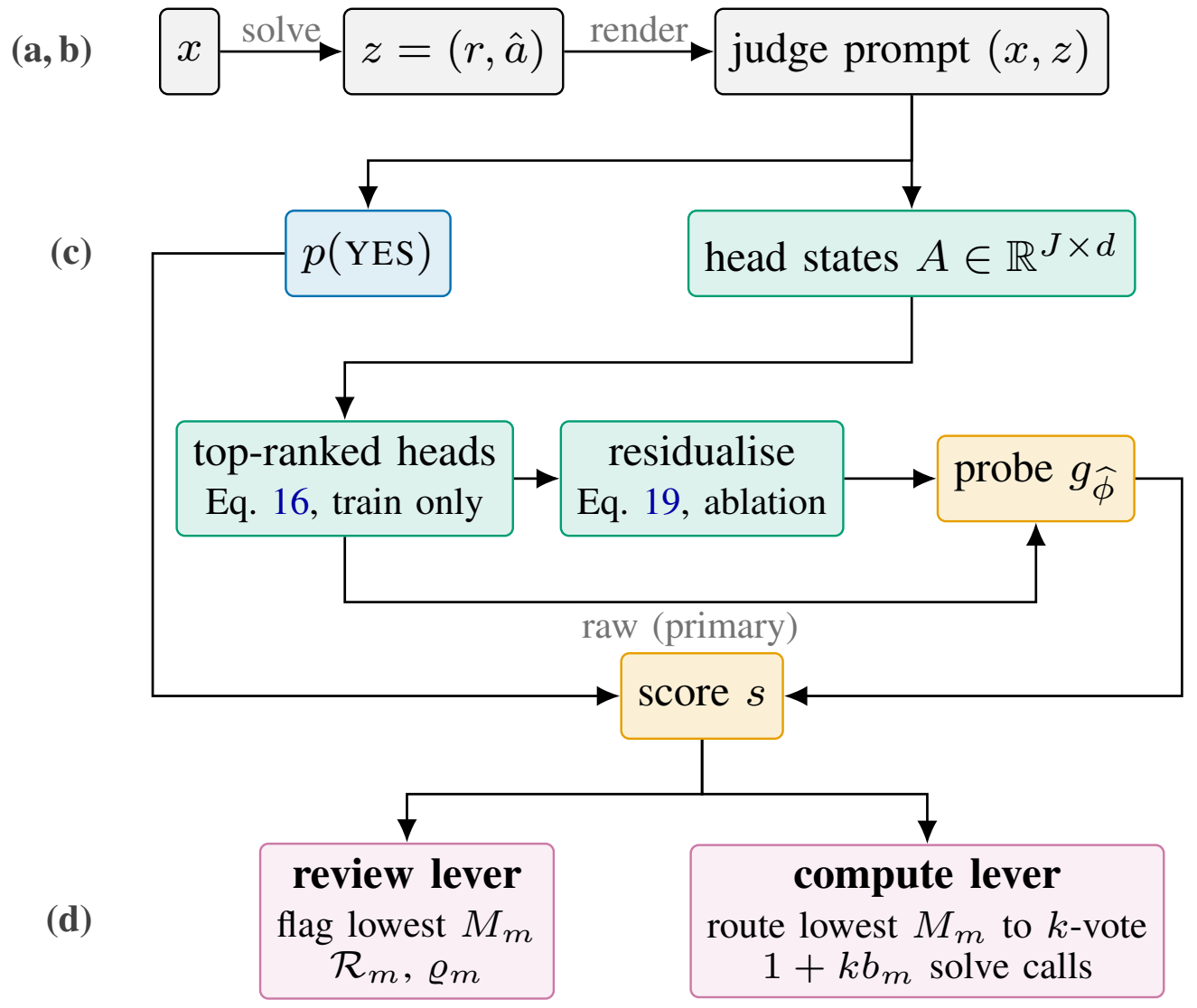
\begin{figure}[t]
  \centering
  \resizebox{\columnwidth}{!}{\input{figures/pipeline}}
  \caption{The control. \textbf{(a)}~The model solves $\query$, producing
  $\sol=(r,\pred)$. \textbf{(b)}~$(\query,\sol)$ is rendered into a judge prompt;
  the black-box baseline reads the \yes/\no\ token mass at the verdict position
  (\cref{eq:pyes}). \textbf{(c)}~The same prompt is replayed and the pre-projection
  attention-head state is recorded; the heads with the largest train-only Fisher
  separability (\cref{eq:alpha}) are concatenated and, in the \emph{controlled}
  ablation, residualised against shallow text statistics (\cref{eq:residual-fit}).
  \textbf{(d)}~A probe maps the result to a correctness score, whose \emph{ranking}
  drives either the review lever or the compute lever of
  \cref{sec:setting-levers}.}
  \label{fig:pipeline}
\end{figure}

\subsection{Black-box baseline: the judge's token mass}
\label{sec:method-baseline}

The baseline this paper compares against is not the verbalized verdict string but
the distribution it was sampled from. Rendering $(\query,\sol)$ into the judge
prompt and writing $q(\cdot)$ for the model's next-token distribution at the
verdict position, with $V_{\yes}$ and $V_{\no}$ the token sets that realise each
verdict,
\begin{equation}
  \label{eq:pyes}
  \pyes(\query,\sol) \;=\;
  \frac{\sum_{v\in V_{\yes}} q(v)}
       {\sum_{v\in V_{\yes}} q(v) + \sum_{v\in V_{\no}} q(v)} .
\end{equation}
This is the principal black-box comparator studied here: it needs only output access, it is
continuous, and it is what an API that returns log-probabilities exposes. The
sampled binary verdict $v\in\{\yes,\no\}$ is a thresholded, one-bit
degradation of \cref{eq:pyes}; its AUROC is mathematically valid but contains
many ties and discards most of the distribution's ranking information, so it is
not an operational baseline in the main results. \cref{app:pyes} audits
verdict-token matching and
the top-$20$ log-probability cap exposed by a commercial API.

\subsection{Reading attention-head activations}
\label{sec:method-readout}

To obtain white-box features, we replay the exact judge prompt with the same model
checkpoint. At the final prompt position, we record each layer's attention output
immediately before the output projection. This vector contains $H$ separate head
outputs, each of width $d$. Keeping these blocks separate lets us select individual
heads; the residual stream would mix their contributions. Stacking all heads from
the $L$ layers gives
\begin{equation}
  \label{eq:activations}
  \acts \in \Real^{N\times J\times d}, \qquad J = L\cdot H .
\end{equation}
The extractor also verifies that the checkpoint, tokenizer, prompt, and assistant
prefill match those used to produce the recorded verdict. It stops on a mismatch,
which could otherwise shift the token position and silently associate the wrong
activation with an answer (\cref{app:repro}).

\subsection{Head selection}
\label{sec:method-heads}

With $J$ in the thousands, retaining every head gives a $Jd$-dimensional feature
far exceeding $N_{\mathrm{tr}}$. We therefore rank heads using a Fisher-based
statistic computed on the training split. Let $\mu_j^{c}$ be the class mean of
head $j$ under $y=c$ and
\begin{equation}
  \label{eq:cov}
  \Sigma_j = \frac{1}{N_{\mathrm{tr}}-2}
  \sum_{i\in\Dtr}\big(\acts_{i,j}-\mu_j^{y_i}\big)\big(\acts_{i,j}-\mu_j^{y_i}\big)^{\!\top}
  + \lambda I
\end{equation}
its ridge-regularised within-class covariance. The Fisher direction
$w_j=\Sigma_j^{-1}(\mu_j^{1}-\mu_j^{0})$ maximises the between- to within-class
scatter ratio on that head, giving the orientation-invariant separability score
\begin{equation}
  \label{eq:alpha}
  \headscore_j = \max\!\left\{
  \begin{aligned}
    &\auroc\big(y,\acts_{:,j}w_j\big),\\[-2pt]
    &1-\auroc\big(y,\acts_{:,j}w_j\big)
  \end{aligned}
  \right\}.
\end{equation}
Let $n_{\mathrm h}$ be the number of heads retained, and let $\heads$ contain
the $n_{\mathrm h}$ heads with the largest values of $\headscore_j$. For answer
$i$, we concatenate their activations:
\begin{equation}
  \label{eq:features}
  |\heads|=n_{\mathrm h}, \qquad
  \feats_i = \big[\acts_{i,j}\big]_{j\in\heads}
  \in \Real^{n_{\mathrm h}d}.
\end{equation}

Each score is computed only on the training split and is used to rank heads.
Under a shared-covariance Gaussian model, the population Fisher direction has
AUROC $\Phi(\delta_j/\sqrt{2})$, so its ranking follows Mahalanobis class
separation (\cref{prop:fisher}). This is a population explanation, not a claim
that the model holds exactly for our regularised finite-sample scores. Choosing
$n_{\mathrm h}$ reduces the feature dimension; it does not assert that
correctness information is sparse across heads. Indeed, the linear probe
continues to improve as $n_{\mathrm h}$ increases in \cref{app:sweep}.

\subsection{Probe}
\label{sec:method-probe}

The probe maps the selected-head representation to a correctness score. The
primary probe uses the raw feature $\feats_i$ from \cref{eq:features}; the
controlled ablation introduced next uses $\resfeats_i$. Let $U_i$ denote the
representation supplied to the classifier. Each coordinate of $U_i$ is
standardised using statistics from $\Dtr$, and
\begin{equation}
  \label{eq:probe}
  \score(\query_i,\sol_i) = g_{\phihat}(U_i) \in [0,1],
\end{equation}
where a larger score indicates stronger evidence that the answer is correct. We
evaluate two classifiers: $\ell_2$-regularised logistic regression with balanced
class weights and a one-hidden-layer MLP. MLP training depends on its random
initialisation, so we fit $n_{\mathrm{seed}}$ models and average their outputs.
This averaged score, rather than any individual fit, determines every ranking,
threshold, curve, and table entry.

\paragraph*{Frozen primary configuration.}
To avoid selecting a method based on evaluation performance, we use one primary
configuration throughout: raw features from $n_{\mathrm h}=32$ selected heads
and an MLP ensemble with $n_{\mathrm{seed}}=5$. This configuration is unchanged
across models, datasets, and budgets. The controlled probe, shallow-feature
detector, and logistic regression are reported only as ablations; none is selected
per evaluation cell or operating point.

\subsection{Shortcut control and its sensitivity to domain shift}
\label{sec:method-control}

A probe could score well by reading how a solution \emph{looks} rather than
whether it is correct. We therefore record seven surface features for each
solution: its character, word, line, digit, and arithmetic-symbol counts; whether
it contains the \texttt{\#\#\#\#} answer marker; and the log-transformed
magnitude of its extracted numeric answer. We standardise these features using
$\Dtr$ and place them, together with an intercept, in the matrix $X$.

On $\Dtr$, we fit a multivariate least-squares map $\Bhat$ that predicts the
selected-head features $\feats$ from $X$. The controlled features are what this
map does not predict:
\begin{equation}
  \label{eq:residual-fit}
  \resfeats_{\mathrm{tr}}=\feats_{\mathrm{tr}}-X_{\mathrm{tr}}\Bhat,
  \qquad
  \resfeats_{\mathrm{te}}=\feats_{\mathrm{te}}-X_{\mathrm{te}}\Bhat.
\end{equation}
On the training split, $\resfeats$ is linearly uncorrelated with the seven
measured features. This is only a limited control: nonlinear or unmeasured surface
cues may remain, while useful semantic information correlated with the measured
features may also be removed.

The same fitted map is used at test time. If the relation between surface features
and head activations changes across domains, this residualisation introduces an
additional shift term, formalised next.

\begin{lemma}[Source-fitted residualisation under domain shift]
  \label{prop:residual-shift}
  Let $B_{\srcdom}$ and $B_{\tgtdom}$ be the population least-squares maps from
  the surface features $X$ to the head features $\feats$ in the source and target
  domains. If the source map is applied to target data, the resulting residual
  $\resfeats_{\tgtdom}:=\feats_{\tgtdom}-X_{\tgtdom}B_{\srcdom}$ satisfies
  \begin{equation}
    \label{eq:shift-decomp}
    \resfeats_{\tgtdom}
    = \resfeats^{\star}_{\tgtdom}+X_{\tgtdom}\Delta B,
    \qquad
    \Delta B:=B_{\tgtdom}-B_{\srcdom},
  \end{equation}
  where
  $\resfeats^{\star}_{\tgtdom}:=\feats_{\tgtdom}-X_{\tgtdom}B_{\tgtdom}$ is
  the residual obtained using the target-domain map.
\end{lemma}

The identity (proved in \cref{app:proof-shift}) shows what the controlled probe
adds under domain shift: the term $X_{\tgtdom}\Delta B$, which is absent from the
raw features. This is a population explanation, not a performance guarantee.
\Cref{sec:trust-boundary} reports an observed pattern consistent with this
mechanism.

\subsection{Estimation protocol}
\label{sec:method-protocol}

All learned components --- the head set $\heads$, feature standardisation,
residualiser $\Bhat$, probe parameters $\phihat$, and any threshold --- are fitted
using $\Dtr$. They are fixed before the corresponding configuration is evaluated
on $\Dte$. In particular, head selection uses only $\Dtr$; selecting
$n_{\mathrm h}$ heads from the $J=L\cdot H$ candidates using test performance
would bias the reported AUROC upward.

We assess test-item and fitting variation separately. Paired bootstrap intervals
resample the held-out test items while keeping the generated answers and fitted
ensemble fixed. Separate analyses then measure variation across MLP
initialisations and training resamples, including changes in the selected heads
(\cref{sec:setup-stats,sec:limitations}).

%% file: figures/pipeline.tex
\begin{tikzpicture}[
    font=\footnotesize,
    box/.style={draw, rounded corners=1.5pt, align=center, inner xsep=3pt,
                inner ysep=2.5pt, line width=0.4pt, minimum height=5.5mm},
    gen/.style ={box, fill=black!5},
    bb/.style  ={box, fill=oiblue!12,   draw=oiblue},
    wb/.style  ={box, fill=oigreen!13,  draw=oigreen},
    pr/.style  ={box, fill=oiorange!16, draw=oiorange},
    lev/.style ={box, fill=oipurple!12, draw=oipurple, align=center},
    flow/.style={-{Latex[length=1.5mm,width=1.3mm]}, line width=0.45pt},
    flowlabel/.style={font=\scriptsize, text=black!55, inner sep=0pt},
    stage/.style={font=\scriptsize\bfseries, text=black!75},
    note/.style={font=\scriptsize, text=black!55, inner sep=1.5pt,
             fill=white, rounded corners=1pt},
  ]

  \node[gen] (q)   at (0,0)      {$\query$};
  \node[gen] (sol) at (1.7,0)    {$\sol=(r,\pred)$};
  \node[gen] (jp)  at (4.65,0)   {judge prompt $(\query,\sol)$};
  \draw[flow] (q)   -- node[midway, above=1.5pt, flowlabel] {solve} (sol);
  \draw[flow] (sol) -- node[midway, above=1.5pt, flowlabel] {render} (jp);
  \node[stage, anchor=east] at (-0.5,0) {(a,\,b)};

  \node[bb] (pyes)  at (1.15,-1.3) {$\pyes$};
  \node[wb] (acts)  at (4.65,-1.3) {head states $\acts\in\Real^{J\times d}$};
  \draw[flow] (jp.south) -- ++(0,-0.42) -| (pyes.north);
  \draw[flow] (jp.south) -- ++(0,-0.42) -| (acts.north);
  \node[stage, anchor=east] at (-0.5,-1.3) {(c)};

  \node[wb] (sel) at (1.0,-2.75)  {top-ranked heads\\\scriptsize Eq.~\ref{eq:alpha}, train only};
  \node[wb] (res) at (3.3,-2.75)  {residualise\\\scriptsize Eq.~\ref{eq:residual-fit}, ablation};
  \node[pr] (g)   at (5.45,-2.75) {probe $g_{\phihat}$};
  \draw[flow] (acts.south) -- ++(0,-0.42) -| (sel.north);
  \draw[flow] (sel) -- (res);
  \draw[flow] (res) -- (g);
  \draw[flow] (sel.south) -- ++(0,-0.42)
        -| node[pos=0.25, below, note] {raw (primary)} (g.south);

  \node[pr] (s) at (3.3,-4.15) {score $\score$};
  \draw[flow] (g.east) -- ++(0.3,0) |- (s.east);
  \draw[flow] (pyes.west) -- ++(-0.85,0) |- (s.west);

  \node[lev] (rev) at (1.4,-5.6)
       {\textbf{review lever}\\\scriptsize flag lowest $M_\budget$\\[-1.5pt]
        \scriptsize $\recall_{\budget}$, $\resid_{\budget}$};
  \node[lev] (rt)  at (4.75,-5.6)
       {\textbf{compute lever}\\\scriptsize route lowest $M_\budget$ to $k$-vote\\[-1.5pt]
        \scriptsize $1+k b_\budget$ solve calls};
  \draw[flow] (s.south) -- ++(0,-0.35) -| (rev.north);
  \draw[flow] (s.south) -- ++(0,-0.35) -| (rt.north);
  \node[stage, anchor=east] at (-0.5,-5.6) {(d)};

\end{tikzpicture}

%% file: sections/05-setup.tex
\section{Experimental Setup}
\label{sec:setup}

\subsection{Models and generation}
\label{sec:setup-models}

We evaluate three open instruction-tuned models: Gemma~3~12B~IT,
Qwen2.5~7B~Instruct, and Llama~3.1~8B~Instruct. Each model serves as both solver
and judge, so its black-box score and attention-head probe assess the same
generated answer.

The six main model--dataset cells use greedy decoding. The compute-allocation
experiment adds two separate GSM8K routing arms, for Llama~3.1 and Gemma~3. Each
routed problem receives $k=5$ new samples generated with temperature $0.7$ and
top-$p=0.95$ (\cref{sec:results-routing}). GSM8K generation uses bf16 weights and
a \num{2048}-token reply budget; MATH uses FP8 weights and a \num{4096}-token
reply budget. Cross-dataset transfer therefore evaluates a joint change in
dataset, weight precision, and generation budget rather than a dataset-only shift.

\subsection{Datasets and analysis population}
\label{sec:setup-data}

We use the official train and test splits of GSM8K~\cite{cobbe2021gsm8k} and
MATH~\cite{hendrycks2021math}. All learned components are fitted on the training
split, and results are reported on the held-out test split. Correctness labels use
dataset-specific answer matching: GSM8K compares parsed numeric values, while
MATH normalises \LaTeX{} and accepts numeric equality when both answers are plain
numeric literals. The MATH rule does not recognize every mathematically
equivalent form; the exact rules are given in \cref{app:equivalence}.

We exclude generations that are truncated, have no parseable reference answer,
or end in a generation error because they do not support the same answer-level
correctness evaluation. This excludes $1{,}682$ of $63{,}876$ records. The
reported results therefore describe labelable, non-truncated generations. Per-cell
sample sizes appear in \cref{tab:main}, with complete exclusion counts and their
effect on error prevalence in \cref{app:populations}. Across the six analysed test
cells, error prevalence ranges from $6.1\%$ to $49.4\%$, which is relevant when
comparing prevalence-sensitive quantities such as average precision.

\subsection{Probe configuration}
\label{sec:setup-probe}

The primary configuration is fixed across models, datasets, and budgets: the
$n_{\mathrm h}=32$ heads ranked on the training split, raw selected-head
features, and a one-hidden-layer MLP with $128$ units. We train the MLP from five
random initialisations and average their output scores. All primary results use
this fixed ensemble. The controlled probe, shallow-feature detector, and logistic
regression are evaluated only as ablations and are never substituted for the
primary method based on test performance. \cref{app:sweep} reports sensitivity
to the number of heads,
regularisation, and MLP width. Optimisation details are provided in
\cref{app:repro}.

\subsection{Statistical analysis}
\label{sec:setup-stats}

Differences between detectors use paired 95\% bootstrap intervals over held-out
test items with \num{2000} resamples. Resampling is class-stratified for
discrimination and review metrics and unstratified for routing. Fitted models and
score vectors remain fixed, while the fixed-budget ranking is recomputed within
each resample. Marginal recall intervals use the Wilson method. When a budget
cuts through tied scores, we report the exact expectation under uniform random
tie-breaking (\cref{app:ties}).

These intervals measure variation across test items conditional on the generated
answers and fitted ensemble. Variation across MLP initialisations and training
resamples is reported separately (\cref{app:bootstrap}). Intervals are nominal
and reported per model--dataset cell; no family-wise correction is applied
(\cref{app:multiplicity}).

%% file: sections/06-results.tex
\section{Results}
\label{sec:results}

\input{tables/table1_main}

\subsection{Discrimination}
\label{sec:results-discrimination}

\Cref{tab:main} and \cref{fig:discrimination} report performance over the full ranking.
The probe improves error-class average precision in all six model--dataset cells,
by $0.081$--$0.237$; every paired interval excludes zero. It improves AUROC in
five cells, by $0.030$--$0.169$. The exception is Qwen2.5~7B/GSM8K, where the
$-0.015$ difference $[-0.048,+0.017]$ is not distinguishable from zero.

The comparison depends on judge strength. On GSM8K, the Gemma~3 and Qwen2.5
judges already attain AUROCs of $0.848$ and $0.873$. The probe exceeds the first
by $0.030$ but not the second. Probe AUROC is numerically higher on MATH for all
three models, but the MATH configuration also changes weight precision and the
generation budget. We therefore treat this as a difference between the tested
configurations, not an isolated dataset effect.

\subsection{Fixed review budget}
\label{sec:results-budget}

\begin{figure*}[t]
  \centering
  \includegraphics[width=\textwidth]{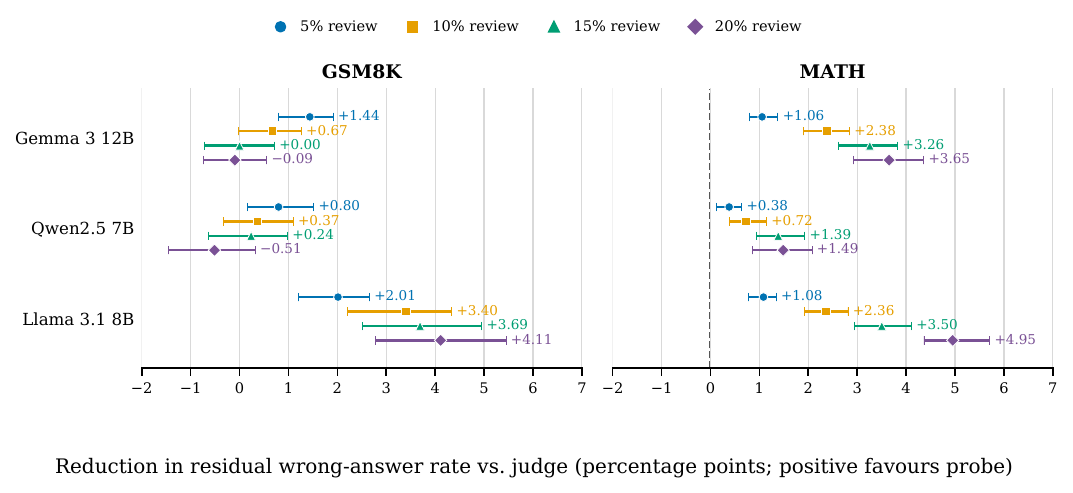}
  \caption{Probe improvement over the judge at matched review budgets. Each point
  is the reduction in the wrong-answer rate among \emph{unreviewed} answers, in
  percentage points; the exact estimate is printed beside its paired 95\%
  class-stratified bootstrap interval (2000 resamples). Positive values favour
  the probe, and intervals crossing zero do not resolve the ordering. The probe
  is fixed across cells and budgets. \Cref{tab:budget} reports the primary 5\%
  operating point, and \cref{prop:residual} links its recall and residual-risk
  views.}
  \label{fig:budget}
\end{figure*}

\input{tables/table2_budget}

Whole-ranking metrics do not determine performance at a specific review
capacity. We therefore give both methods the same budget and flag exactly the
$M_\budget$ lowest-scoring answers. This comparison requires no label-dependent
threshold tuning on the test set.

At the primary 5\% budget, the probe retrieves more errors in all six cells
(\cref{tab:budget}). Recall gains range from $0.013$ to $0.225$, and every paired
interval excludes zero. These gains reduce the wrong-answer rate among
unreviewed answers by $0.38$--$2.01$ percentage points. The largest reduction is
for Llama/GSM8K. Recall gain and residual-risk reduction need not have the same
ordering because \cref{eq:residual-delta} scales the former by the cell's error
prevalence.

The advantage is less uniform beyond this operating point (\cref{fig:budget}).
At 10\% review, all six point estimates favour the probe, but only four paired
intervals exclude zero. On MATH, the probe remains ahead at every displayed
budget. On GSM8K, its advantage grows for Llama but converges to zero for Gemma~3
and Qwen2.5; at 20\%, both of the latter intervals cross zero. This is why we
report the deployment operating point separately from AUROC, which averages over
the full ranking (\cref{prop:auroc}).

\subsection{Compute allocation}
\label{sec:results-routing}

\begin{figure*}[t]
  \centering
  \includegraphics[width=\textwidth]{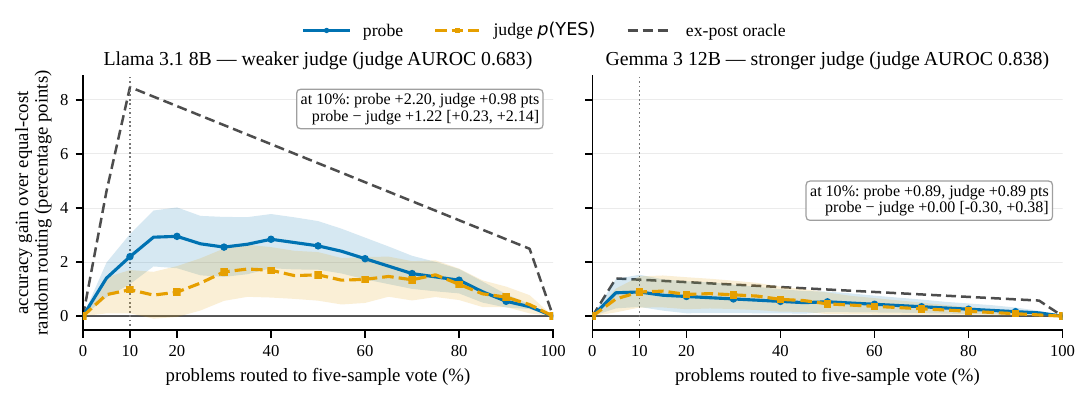}
  \caption{Routing additional inference on GSM8K ($k=5$). Each policy sends the
  indicated fraction of problems to a five-sample majority vote and retains the
  greedy answer elsewhere; routing fraction $\budget$ costs $1+5\budget$ solver
  calls per problem. The vertical axis is accuracy gain over selecting the same
  number of problems uniformly at random (\cref{eq:route-random}); zero is the
  equal-cost random baseline. Shaded regions are paired 95\% bootstrap intervals
  for the probe and judge. Boxes report their exact comparison at the primary
  10\% routing budget. The dashed oracle observes which answers sampling will
  repair or harm and is not deployable. Both panels share a vertical scale,
  exposing Gemma~3's smaller available headroom.}
  \label{fig:routing}
\end{figure*}

Compute allocation asks whether the score can target extra sampling. We evaluate
two $k=5$ GSM8K arms at opposite ends of the judge-quality range: Llama~3.1~8B
has the weakest judge, whereas Gemma~3~12B has one of the strongest. Their
available gains differ substantially. For Llama, five-sample voting raises
accuracy from $82.5\%$ to $89.5\%$, repairing 120 answers and breaking 28
($N=1308$). For Gemma~3, it raises accuracy from $94.2\%$ to $95.1\%$, repairing
19 answers and breaking 7 ($N=1319$). Thus, Gemma~3 offers less headroom before
the routing methods are compared.

\paragraph*{Both arms beat random allocation.}
At $\budget=0.10$, or $1.5$ solve calls per problem, the probe improves accuracy
over equal-cost random routing by $\num{2.20}$ points on Llama (95\% CI
$[\num{1.18},\num{3.04}]$) and $\num{0.89}$ points on Gemma~3
$[\num{0.34},\num{1.53}]$. Although the Gemma~3 gain is smaller in absolute
terms, it captures $66\%$ of the oracle gain available at this budget, compared
with $26\%$ for Llama. Shallow features alone yield $+1.21$
$[+0.31,+1.98]$ and $+0.59$ $[+0.11,+1.09]$ points, respectively, showing that
useful routing signal is also present without activations
(\cref{sec:trust-boundary}).

\paragraph*{Only the weaker-judge arm beats the judge.}
For Llama, the probe exceeds judge-guided routing by $\num{1.22}$ points at
$\budget=0.10$ (paired 95\% CI $[\num{0.23},\num{2.14}]$) and $\num{2.06}$ points
at $\budget=0.20$ $[\num{0.92},\num{3.13}]$. For Gemma~3, probe- and
judge-guided routing attain the same accuracy at $\budget=0.10$: their difference
is $\num{0.00}$ points $[-\num{0.30},\num{0.38}]$. Both outperform random
routing, showing that each identifies answers that benefit from additional
inference. This comparison establishes useful selection relative to an
uninformed baseline, but not that the probe is better than the judge. Across
these two experiments, the probe outperforms the judge only for Llama, whose
judge score is less discriminative.

\paragraph*{The oracle quantifies the remaining headroom.}
The oracle knows in advance whether additional sampling will repair, preserve,
or break each answer. It therefore routes all repairable errors while avoiding
answers that would be harmed. For Llama, all repairs are captured after routing
$9.2\%$ of the problems, yielding $91.7\%$ accuracy. This is $28/1308=2.14$
percentage points higher than routing every problem, because five-sample voting
breaks 28 initially correct answers. For Gemma~3, the oracle reaches $95.6\%$
accuracy after routing $1.4\%$ of the problems and avoids 7 harmful changes, a
$7/1319=0.53$-point advantage over routing every problem
(\cref{eq:route-oracle}). A deployable probe cannot know these outcomes in
advance: its correctness score indicates whether the current answer may be
wrong, not whether additional samples will fix it (\cref{sec:setting-routing}).

These results are illustrative rather than a broad evaluation of routing. They
cover only GSM8K, $k=5$, and one fixed set of five sampled completions per
problem. Results may change with the dataset, sampling budget, or generated
completions. Moreover, \cref{tab:main} evaluates the $k=1$ populations, whereas
the routing experiments use separate $k=5$ populations; their values are
therefore not directly comparable (\cref{sec:limitations}).

%% file: tables/table1_main.tex
\begin{table*}[t]
\centering
\caption{Held-out discrimination, six model--dataset cells. The probe is the
primary method fixed in advance (raw top-32-head five-seed MLP ensemble); the
baseline is the judge's continuous $\pyes$. $\prev$ is the observed error
prevalence in the analysed population. Deltas are probe minus judge with paired
class-stratified 95\% bootstrap intervals over test items (2000 resamples);
the six cells are not family-wise adjusted. AP treats errors as the positive
class and is therefore sensitive to $\prev$. Intervals are conditional on the
generated answer population and fitted ensemble (\cref{sec:limitations}).
Within each probe--judge pair, bold marks the higher point estimate; the
paired interval indicates whether the difference is statistically resolved.}
\label{tab:main}
\small
\begin{tabular}{llrrrrrrrr}
\toprule
& & & & \multicolumn{3}{c}{AUROC (correctness)} & \multicolumn{3}{c}{AP (error class)} \\
\cmidrule(lr){5-7}\cmidrule(lr){8-10}
Model & Data & $N$ & $\prev$ & probe & judge & $\Delta$ [95\% CI] & probe & judge & $\Delta$ [95\% CI] \\
\midrule
Gemma~3 12B & GSM8K & 1319 & 0.061 & \textbf{0.879} & 0.848 & +0.030~[+0.001, +0.063] & \textbf{0.498} & 0.295 & +0.203~[+0.101, +0.292] \\
Qwen2.5 7B & GSM8K & 1319 & 0.081 & 0.858 & \textbf{0.873} & -0.015~[-0.048, +0.017] & \textbf{0.538} & 0.419 & +0.120~[+0.028, +0.201] \\
Llama~3.1 8B & GSM8K & 1307 & 0.178 & \textbf{0.813} & 0.697 & +0.116~[+0.079, +0.157] & \textbf{0.584} & 0.350 & +0.233~[+0.169, +0.296] \\
\midrule
Gemma~3 12B & MATH & 4883 & 0.184 & \textbf{0.914} & 0.822 & +0.092~[+0.077, +0.107] & \textbf{0.776} & 0.539 & +0.237~[+0.205, +0.267] \\
Qwen2.5 7B & MATH & 4949 & 0.270 & \textbf{0.888} & 0.848 & +0.040~[+0.030, +0.050] & \textbf{0.771} & 0.690 & +0.081~[+0.061, +0.103] \\
Llama~3.1 8B & MATH & 4466 & 0.494$^{\dagger}$ & \textbf{0.865} & 0.696 & +0.169~[+0.154, +0.184] & \textbf{0.867} & 0.673 & +0.194~[+0.176, +0.212] \\
\addlinespace[2pt]
\multicolumn{10}{p{0.96\textwidth}}{\footnotesize $^{\dagger}$Llama~3.1~8B error prevalence rises from 0.178 on GSM8K to 0.494 on MATH. Because dataset, weight precision, and reply budget change together, this is an observed population shift, not an isolated dataset effect.} \\
\bottomrule
\end{tabular}
\end{table*}

%% file: tables/table2_budget.tex
\begin{table*}[t]
\centering
\caption{Primary matched 5\% review operating point. Both methods flag the
same number of answers. Error recall is the fraction of wrong answers
flagged. $\Delta$ recall is probe minus judge; unreviewed error-rate
reduction is the corresponding decrease among answers not flagged. The
final-column brackets are paired class-stratified 95\% bootstrap intervals
(2000 resamples). Positive values favour the probe. Bold marks the higher
recall point estimate; the paired interval quantifies its uncertainty.}
\label{tab:budget}
\small
\setlength{\tabcolsep}{6pt}
\begin{tabular}{llrrrr}
\toprule
& & \multicolumn{2}{c}{Error recall at 5\% review} &
\multicolumn{2}{c}{Probe advantage over judge} \\
\cmidrule(lr){3-4}\cmidrule(lr){5-6}
Model & Data & probe & judge & $\Delta$ recall &
unreviewed error-rate reduction (pp) [95\% CI] \\
\midrule
Gemma~3 12B & GSM8K & \textbf{0.487} & 0.263 & +0.225 & +1.44~[+0.80, +1.92] \\
Qwen2.5 7B & GSM8K & \textbf{0.402} & 0.308 & +0.093 & +0.80~[+0.16, +1.52] \\
Llama~3.1 8B & GSM8K & \textbf{0.227} & 0.120 & +0.107 & +2.01~[+1.21, +2.66] \\
\midrule
Gemma~3 12B & MATH & \textbf{0.254} & 0.200 & +0.055 & +1.06~[+0.80, +1.38] \\
Qwen2.5 7B & MATH & \textbf{0.170} & 0.157 & +0.013 & +0.38~[+0.13, +0.64] \\
Llama~3.1 8B & MATH & \textbf{0.101} & 0.080 & +0.021 & +1.08~[+0.78, +1.34] \\
\bottomrule
\end{tabular}
\end{table*}

%% file: sections/07-trust-boundary.tex
\section{Trust Boundary Under Protocol Shift}
\label{sec:trust-boundary}

\input{tables/table3_transfer}

The evidence in \cref{sec:results} assumes that fitting and deployment use the
same dataset and inference protocol. We now test the boundary of that claim by
fitting on one mathematics dataset and scoring the other. This transfer also
changes weight precision and reply budget, so it evaluates a joint deployment
shift rather than dataset shift alone.

\subsection{The primary probe does not transfer reliably}
\label{sec:transfer}

The primary probe loses AUROC in all six transfer directions. Relative to
in-domain fitting, the paired changes range from $-0.040$
$[-0.080,-0.001]$ to $-0.270$ $[-0.289,-0.251]$; every interval excludes zero
(\cref{tab:transfer} and \cref{tab:transfer-full}). Thus, the in-domain ranking does not
persist under any evaluated joint shift.

The relevant deployment baseline is the target-domain judge score $\pyes$,
which requires no probe fitting. It outperforms the transferred primary probe in
five of six directions. The sole exception is Llama~3.1~8B transferred from MATH
to GSM8K, where the probe leads by $0.060$ AUROC
$[+0.023,+0.099]$. This isolated win does not support transferring the probe
without target-domain validation.

The attribution must remain narrow. GSM8K uses bf16 weights and a 2048-token
reply limit, whereas MATH uses FP8 weights and 4096 tokens
(\cref{sec:setup-models}). Each transfer direction therefore changes dataset,
numeric precision, and generation budget together. The experiment establishes
failure under that combined change, but cannot isolate its cause.

\subsection{Why residualisation may amplify the shift}
\label{sec:residualiser}

Residualisation introduces a specific source of sensitivity. By
\cref{prop:residual-shift}, applying a source-domain residualiser to target data
adds $X_{\tgtdom}\Delta B$ whenever the relation between surface features and
activations changes across domains. This identity does not determine whether the
term helps or hurts, but it identifies a mechanism absent from the raw features.

The GSM8K$\to$MATH results are consistent with amplification. The controlled
probe loses more AUROC than the raw probe for Gemma~3 ($0.335$ versus $0.106$),
Qwen2.5 ($0.249$ versus $0.125$), and Llama~3.1 ($0.405$ versus $0.270$) on the
same target items (\cref{tab:transfer-full}). For Llama, the controlled AUROC
falls to $0.448$, below chance and therefore indicative of an inverted ranking.

This pattern is not a causal test of the decomposition. The raw and controlled
probes are fitted separately, and the intervals quantify their individual
transfer losses rather than the difference between those losses. A direct test
would hold precision and reply budget fixed, compare the paired loss gap, and
evaluate a residualiser refitted on target-domain data. The present evidence
therefore supports only that residualisation \emph{may} exacerbate this shift.

\subsection{Shallow features are stable but not uniformly better}
\label{sec:shallow-fallback}

The seven-feature detector has the smallest transfer-loss point estimate in all
six directions, with a maximum loss of $0.030$ AUROC. Three intervals include
zero: both Gemma~3 directions and Qwen2.5 MATH$\to$GSM8K
(\cref{tab:transfer-full}). The clearest case is Gemma~3
GSM8K$\to$MATH: the shallow detector changes by only $-0.0001$
$[-0.005,+0.005]$, while the raw and controlled probes lose $0.106$ and $0.335$.

Stability alone does not imply better discrimination. The shallow detector
loses to the target-domain judge in four directions, is unresolved in one, and
wins only for Llama~3.1 GSM8K$\to$MATH, by $0.075$ AUROC
$[+0.057,+0.094]$. In that same direction, the transferred primary probe loses
to the judge by $0.100$ $[-0.117,-0.084]$. Shallow features are therefore a
useful output-only reference, but the evidence does not justify treating them as
a universal fallback.

\subsection{Deployment implication}
\label{sec:deployment-reading}

In-domain performance does not justify deploying a fitted probe unchanged after
the data or inference protocol changes. The probe should be revalidated on
labelled data from the intended model--dataset--generation configuration. When
such validation is unavailable, the target-domain judge is the conservative
fallback supported by these experiments: it needs no probe fitting and
outperforms the transferred primary probe in five of six directions. This is a
protocol-specific recommendation, not a guarantee for other domains or shifts.

%% file: tables/table3_transfer.tex
\begin{table*}[t]
\centering
\caption{Held-out target-split AUROC under the joint dataset and inference-protocol
change. The transferred probe is fitted on the other dataset; the in-domain probe
is fitted on the target's training split. The judge is the target-domain $\pyes$
and requires no probe fitting. Bold marks the higher value in the operational
comparison between the transferred primary probe and the judge. Deltas are paired
differences on the same target items. All twelve primary-probe 95\% bootstrap
intervals exclude zero; \cref{tab:transfer-full} reports the intervals and the full
ablation matrix. Dataset, precision, and reply budget change together, so their
effects cannot be separated.}
\label{tab:transfer}
\small
\setlength{\tabcolsep}{5pt}
\begin{tabular}{llrrrrrrr}
\toprule
& & \multicolumn{3}{c}{deployment comparison} & \multicolumn{2}{c}{primary-probe transfer} & \multicolumn{2}{c}{transferred ablations} \\
\cmidrule(lr){3-5}\cmidrule(lr){6-7}\cmidrule(lr){8-9}
Model & Target & judge & probe & $\Delta$ vs.\ judge & in-domain & $\Delta$ vs.\ in-dom. & controlled & shallow \\
\midrule
Gemma~3 12B & GSM8K & \textbf{0.848} & 0.808 & -0.040 & 0.879 & -0.071 & 0.802 & 0.738 \\
 & MATH & \textbf{0.822} & 0.808 & -0.014 & 0.914 & -0.106 & 0.567 & 0.801 \\
Qwen2.5 7B & GSM8K & \textbf{0.873} & 0.819 & -0.054 & 0.858 & -0.040 & 0.774 & 0.711 \\
 & MATH & \textbf{0.848} & 0.763 & -0.085 & 0.888 & -0.125 & 0.624 & 0.772 \\
Llama~3.1 8B & GSM8K & 0.697 & \textbf{0.757} & +0.060 & 0.813 & -0.055 & 0.723 & 0.692 \\
 & MATH & \textbf{0.696} & 0.595 & -0.100 & 0.865 & -0.270 & 0.448 & 0.771 \\
\bottomrule
\end{tabular}
\end{table*}

%% file: sections/08-limitations.tex
\section{Limitations}
\label{sec:limitations}

\paragraph*{Transfer and external validity}
GSM8K$\leftrightarrow$MATH changes the dataset, weight precision, and reply
budget together. The transfer results therefore establish degradation only
under this joint shift; matched-precision and matched-budget runs are needed to
isolate a dataset effect. The study covers three open 7--12B models and one pair
of mathematics datasets, not frontier models, other domains, or temporal drift.
Routing is narrower still: two models, GSM8K, $k=5$, and one fixed set of sampled
completions.

\paragraph*{Supervision and baseline fairness}
Fitting requires thousands of gold-labelled answers from the same model and
task, whereas $\pyes$ requires no correctness labels. The judge is therefore a
practical fallback, but not the strongest possible supervised output-only
baseline. A fairer access comparison would train the same classifier and split
on $\pyes$, shallow statistics, solver likelihood, and semantic-uncertainty
features.

\paragraph*{Uncertainty and fitting stability}
The six model--dataset cells form a family and the reported per-cell intervals
are nominal and unadjusted. The bootstrap resamples held-out items while keeping
the generated answers, selected heads, and fitted ensemble fixed. The five MLP
initialisations are averaged into one predictor, and train-resample fits are
available only for the controlled ablation. Repeated generation seeds and
full-pipeline resampling are therefore needed for unconditional uncertainty.
Fitting may also be unstable because the probe uses 4096--8192 features with few
minority examples. In addition, oversampling precedes the MLP's internal
validation split, so duplicated minority rows may occur in both fitting and
validation data. The held-out test set remains untouched.

\paragraph*{Labels, exclusions, and run populations}
MATH uses normalised-string equality and does not equate every mathematically
equivalent form, such as $1/2$ and $0.5$; no symbolic-equivalence or manual-audit
sensitivity is available. Truncated generations are excluded, which lowers
observed error prevalence but need not change the probe--judge comparison
conservatively. Counting non-termination as failure would provide a complementary
system-level analysis. The $k=1$ and $k=5$ experiments use separate generated
populations and their accuracies must not be combined (\cref{app:populations}).

\paragraph*{Interpretation of the probe}
Because activations are read from a self-judging pass, a correctness-labelled
probe may track the model's own belief rather than objective correctness when the
two disagree~\cite{lu2026diagnosing}. We do not evaluate the conflict sets needed
to separate these explanations. The paper therefore claims predictive
association, not semantic identification of an objective-correctness direction.

\paragraph*{Contamination is bounded, not excluded}
The GSM-Symbolic comparison uses only 100 templates. Its broad intervals rule out
only large perturbation effects; the full-GSM AUROC comparison is a difference of
independently resampled populations rather than a paired estimate. The templates
derive from GSM8K, so absence from pretraining is not established
(\cref{app:symbolic}).

%% file: sections/09-future-work.tex
\section{Future Work}
\label{sec:future-work}

Two directions follow from our findings. The first is protocol adaptation.
Controlled experiments should vary dataset, weight precision, and generation
budget separately to identify which changes invalidate the probe. They should
then measure whether a small labelled sample from the target setting is sufficient
to recalibrate or refit it. Extending this analysis across model families,
domains, routing budgets, and generation seeds would establish how reliably the
in-domain gains persist.

The second direction is to use internal assurance signals during training, not
only at deployment. For reinforcement learning of self-correcting models, a
repair-aware predictor could estimate the expected change in correctness after
an answer is revised. This signal could prioritize informative training examples
and supplement outcome-based rewards: revisions that repair errors would be
encouraged, while changes that damage correct answers would be discouraged. A
reliable training objective would still require independently verified outcomes
and explicit tests for reward hacking, calibration drift, and transfer. The
present probe offers a representation from which such an objective could be
developed.

%% file: sections/09-conclusion.tex
\section{Conclusion}
\label{sec:conclusion}

Assurance value is determined at the deployment operating point, not by aggregate
discrimination alone. Across six in-domain settings, the hidden-state probe
increases error recall over verbalized self-judgment by
$\num{1.3}$--$\num{22.5}$ percentage points at 5\% review. In two GSM8K routing
experiments, it also improves accuracy over equal-cost random allocation and,
when the judge is weaker, over judge-guided routing. These results show that
internal signals can add operational value beyond verbalized self-judgment. They
also separate two objectives: identifying a likely error and predicting whether
additional inference will repair it.

The transfer experiments define how this capability should be used. Probe AUROC
falls by $0.040$--$0.270$ under the evaluated joint shifts of dataset, precision,
and generation budget, so a score fitted in one configuration should be
revalidated before deployment in another. Within this boundary, hidden-state
probes provide a practical, supervised way to direct scarce review and inference
compute toward the answers most likely to benefit. More broadly, our results
motivate assurance evaluations that pair aggregate ranking metrics with the
actual operating budget and explicitly validate each intended deployment
protocol. Used this way, the probe is evidence for a budgeted assurance control,
not a certification of answer correctness or regulatory compliance.

%% file: appendix/A-proofs.tex
\section{Proofs and Numerical Verification}
\label{app:proofs}

The appendix first proves five results and then checks the three finite-sample
identities against the recorded artifacts. Their scopes are:

\begin{center}\small
\setlength{\tabcolsep}{3pt}
\begin{tabular}{@{}p{0.23\columnwidth}p{0.45\columnwidth}p{0.23\columnwidth}@{}}
\toprule
Result & Type of claim & Where used \\
\midrule
Residual risk (\cref*{prop:residual})
  & Exact finite-sample identity
  & \cref*{sec:setting-review}\newline \cref*{sec:results-budget} \\
AUROC (\cref*{prop:auroc})
  & Population identity with exact finite-sample analogue
  & \cref*{sec:setting-auroc}\newline \cref*{sec:results} \\
Fisher direction (\cref{prop:fisher})
  & Model-dependent population result (shared-covariance Gaussian)
  & \cref*{sec:method-heads} \\
Residual shift (\cref*{prop:residual-shift})
  & Exact algebraic decomposition
  & \cref*{sec:method-control}\newline \cref*{sec:trust-boundary} \\
Routing identities
  & Exact finite-sample identities
  & \cref*{sec:setting-routing}\newline \cref*{sec:results-routing} \\
\bottomrule
\end{tabular}
\end{center}

Here, ``exact'' means that the result follows from the cited definitions without
a distributional model. Keeping proofs and artifact checks separate prevents
numerical agreement from being mistaken for proof.

\subsection{Proofs}

\subsubsection{Residual-risk identity (\texorpdfstring{\cref*{prop:residual}}{Proposition 1})}
\label{app:proof-residual}

The proof expresses both metrics through the same quantity: the expected number
of errors selected for review.

\begin{proof}
Work with the finite-sample quantities on a population of $N$ items containing
$W=\prev N$ errors, and let the budget select $M_\budget$ items, with realized
fraction $b_\budget=M_\budget/N$. For a random resolution of any cutoff tie,
let $C$ be the number of selected errors and let
$\overline C:=\E_{\mathrm{tie}}[C]$. Every tie resolution selects exactly
$M_\budget$ items. Hence, by \cref*{eq:recall,eq:residual},
\begin{equation*}
  \recall_\budget=\frac{\overline C}{W},
  \qquad
  \resid_\budget=\frac{W-\overline C}{N-M_\budget}.
\end{equation*}
Substituting $\overline C=W\recall_\budget$, $W=\prev N$, and
$M_\budget=b_\budget N$ gives
\begin{equation*}
  \resid_{\budget}
  = \frac{W\big(1-\recall_{\budget}\big)}{N-M_\budget}
  = \frac{\prev N \big(1-\recall_{\budget}\big)}{N(1-b_\budget)}
  = \prev\,\frac{1-\recall_{\budget}}{1-b_\budget}.
\end{equation*}
For two scores on the same population and at the same realized budget,
$\prev$ and $b_\budget$ are common. Subtracting the identity for $\score$ from
the identity for $\score'$ therefore gives \cref*{eq:residual-delta}, including
its sign.
\end{proof}
Thus recall and residual risk are not independent measurements: for fixed
$\prev$ and $b_\budget$, either one determines the other exactly. The artifact
check is reported in \cref{app:numerical-residual}.

\subsubsection{AUROC identity (\texorpdfstring{\cref*{prop:auroc}}{Proposition 2})}
\label{app:proof-auroc}

The proof decomposes each increase in flagged population mass into error mass and
correct-answer mass, then integrates error recall along that path.

\begin{proof}
Let $D=-\score$ and treat errors as the positive class, so flagging the
lowest-scoring items is equivalent to thresholding $D$ from above. Along the
randomized threshold path, write $\mathrm{TPR}$ and $\mathrm{FPR}$ for the
error-class and correct-class selection rates, respectively. Uniform
randomization within a tied score group fills in the corresponding line segment
of the ROC curve. At every point on this path, the flagged population fraction
is, by the law of total probability,
\begin{equation}
  \label{eq:budget-decomp}
  \budget
  = \pi_{0,P}\,\mathrm{TPR}
    +(1-\pi_{0,P})\,\mathrm{FPR}.
\end{equation}
As the path moves from selecting no item to selecting every item, $\budget$,
$\mathrm{TPR}$, and $\mathrm{FPR}$ each move monotonically from $0$ to $1$.
Thus \cref{eq:budget-decomp} parameterizes the randomized ROC curve by budget,
and, in the Riemann--Stieltjes sense,
$\mathrm{d}\budget=\pi_{0,P}\,\mathrm{d}\mathrm{TPR}
+(1-\pi_{0,P})\,\mathrm{d}\mathrm{FPR}$. Since
$\recall^P_\budget=\mathrm{TPR}$ along this path,
\begin{align*}
  \int_0^1 \recall^P_{\budget}\,\mathrm{d}\budget
  &= \pi_{0,P}\!\int_0^1 \mathrm{TPR}\,\mathrm{d}\mathrm{TPR}
   \notag\\[-2pt]
  &\quad +(1-\pi_{0,P})\!\int_0^1
        \mathrm{TPR}\,\mathrm{d}\mathrm{FPR} \\
  &= \frac{\pi_{0,P}}{2}
    +(1-\pi_{0,P})\auroc_P.
\end{align*}
The first integral is $\tfrac12$ because $\mathrm{TPR}$ runs from $0$ to $1$.
The second is the ROC area of $D$ when errors are positive. With half credit for
ties, this equals the ROC area of $\score$ when correctness is positive:
\begin{align*}
  &\Prob(D_0>D_1)+\tfrac12\Prob(D_0=D_1) \\
  &\qquad=\Prob(\score_1>\score_0)
    +\tfrac12\Prob(\score_1=\score_0),
\end{align*}
where $\score_0$ and $\score_1$ are the scores of independent error-class and
correct-class draws, respectively, and $D_c=-\score_c$. This is
$\auroc_P(\score)$ and establishes \cref*{eq:auroc-identity}.
\end{proof}

The finite-sample correction is also exact. Assume $0<W<N$, randomly order items
within each tied score group, and let $r_e$ be the resulting rank of error item
$e$ in ascending score order. The item contributes to the captured-error count
at every budget $j/N$ for which $j\ge r_e$. Therefore
\begin{equation*}
  \frac{1}{N}\sum_{j=1}^{N}\recall_{j/N}
  =\frac{1}{NW}\sum_{e:y_e=0}
    \E_{\mathrm{tie}}[N-r_e+1].
\end{equation*}
For every error--correct pair, the probability that the correct item follows the
error item is precisely its AUROC contribution, including half credit for a tie.
These pairs therefore contribute $W(N-W)\auroc$ in expectation. Every unordered
pair of errors contributes one later error, giving $W(W-1)/2$, and each error
also contributes itself, giving $W$. Dividing the total by $NW$ yields
\begin{equation}
  \frac{1}{N}\sum_{j=1}^{N}\recall_{j/N}
  = \frac{\prev}{2}+(1-\prev)\auroc+\frac{1}{2N}.
  \label{eq:auroc-finite}
\end{equation}
The $1/(2N)$ term is the finite-grid endpoint correction; it vanishes as
$N\to\infty$ and is absent from the population integral in
\cref*{eq:auroc-identity}.

The corresponding artifact check is reported in \cref{app:numerical-auroc}.

\subsubsection{Population Fisher direction (\texorpdfstring{\cref{prop:fisher}}{Proposition 3})}
\label{app:proof-fisher}

This proposition explains when the Fisher criterion ranks heads by a population
separation parameter. It does not assert that the Gaussian model holds for the
observed activations.

\begin{proposition}[Population Fisher direction under a shared Gaussian covariance]
  \label{prop:fisher}
  For $c\in\{0,1\}$, let
  $\acts_{:,j}\mid y=c\sim\mathcal N(\mu_j^c,C_j)$, where $C_j$ is a
  positive-definite population covariance shared by the two classes. Let
  $d_j=\mu_j^1-\mu_j^0$ and the unregularised population Fisher direction
  $v_j=C_j^{-1}d_j$. If $\Phi$ is the standard normal cumulative distribution
  function and $\delta_j=(d_j^{\!\top}C_j^{-1}d_j)^{1/2}$, then
  \begin{equation}
    \label{eq:fisher-auc}
    \auroc\big(y,\acts_{:,j}v_j\big)
    =\Phi\!\left(\frac{\delta_j}{\sqrt{2}}\right).
  \end{equation}
\end{proposition}

\begin{proof}
Let $U_c=A_c^{\!\top}v_j$, where
$A_c\sim\mathcal N(\mu_j^c,C_j)$ are independent class-conditional draws. In
both classes, the projected variance is
$v_j^{\!\top}C_jv_j=\delta_j^2$, while the mean difference is
$v_j^{\!\top}d_j=\delta_j^2$.

First consider the degenerate separation $\delta_j=0$. Positive definiteness of
$C_j$ implies $d_j=0$ and hence $v_j=0$. The projected score is constant, so its
AUROC is $1/2$ under the half-credit tie convention; this equals $\Phi(0)$.

Now suppose $\delta_j>0$. Independence gives
$U_1-U_0\sim\mathcal{N}(\delta_j^2,2\delta_j^2)$. Because Gaussian draws tie
with probability zero, the pairwise definition of AUROC gives
\begin{equation*}
  \auroc = \Prob(U_1-U_0>0)
         = \Phi\!\left(\frac{\delta_j^{2}}{\sqrt{2}\,\delta_j}\right)
         = \Phi\!\left(\frac{\delta_j}{\sqrt{2}}\right).
\end{equation*}
Together, the two cases prove \cref{eq:fisher-auc}. Since $\Phi$ is strictly
increasing, population AUROC is strictly increasing in the nonnegative
Mahalanobis separation $\delta_j$.
\end{proof}

The simplification in \cref{eq:fisher-auc} does not apply exactly to the
implemented regularised $w_j=(\widehat C_j+\lambda I)^{-1}\widehat d_j$ with
$\lambda=10^{-2}$; \cref*{eq:alpha} is an empirical train-only selection score.

\subsubsection{Residual shift (\texorpdfstring{\cref*{prop:residual-shift}}{Lemma 1})}
\label{app:proof-shift}

\begin{proof}
Start from the residual formed on target-domain data using the source-domain map.
Adding and subtracting $X_{\tgtdom}B_{\tgtdom}$ gives
\begin{equation*}
  \resfeats_{\tgtdom}
  = \feats_{\tgtdom} - X_{\tgtdom}B_{\srcdom}
  = \underbrace{\big(\feats_{\tgtdom}-X_{\tgtdom}B_{\tgtdom}\big)}_{\resfeats^{\star}_{\tgtdom}}
    + X_{\tgtdom}\big(B_{\tgtdom}-B_{\srcdom}\big).
\end{equation*}
The parenthesized term is exactly the target-map residual
$\resfeats^{\star}_{\tgtdom}$, and the remaining term is
$X_{\tgtdom}\Delta B$. This is \cref*{eq:shift-decomp}. No distributional
assumption is needed for the algebraic identity once the two maps are defined.
\end{proof}

Two qualifications delimit the result. First, it is a decomposition, not a bound
on excess risk: the added term can reinforce or cancel other target-domain shift,
so the lemma alone predicts neither the sign nor magnitude of a performance
change. Second, $\Delta B$ is a population object; on finite samples the fitted
$\Bhat$ also carries estimation error. In GSM8K$\to$MATH the primary MLP's
observed raw-minus-controlled target gaps are $0.139$--$0.241$ AUROC, consistent
with the mechanism, but the experiment does not provide a paired
difference-of-differences interval or isolate residualisation causally.

\subsubsection{Routing comparators}
\label{app:proof-routing}

These identities condition on the realized greedy and majority-vote answers.
They describe how a routing set aggregates the per-item repair values; they do not
assume that a deployable policy can observe those values before routing.

\begin{proof}[Proof of \cref*{eq:route-acc,eq:route-random,eq:route-oracle}]
Define the correctness indicators
\begin{align*}
  a_i^{(1)}&:=\indic{\pred_i^{(1)}\equiv\gold(\query_i)}, \\
  a_i^{(k)}&:=\indic{\pred_i^{(k)}\equiv\gold(\query_i)},
\end{align*}
so $\repair_i=a_i^{(k)}-a_i^{(1)}$.

\emph{Routing identity.}
Under routing set $\routeset$, the released answer is the greedy answer outside
$\routeset$ and the majority-vote answer inside it. Therefore
\begin{align*}
  \acc(\routeset)
  &=\frac{1}{N}\left(\sum_{i\notin\routeset}a_i^{(1)}
      +\sum_{i\in\routeset}a_i^{(k)}\right) \\
  &=\frac{1}{N}\sum_{i=1}^{N}a_i^{(1)}
    +\frac{1}{N}\sum_{i\in\routeset}
      \big(a_i^{(k)}-a_i^{(1)}\big),
\end{align*}
which is \cref*{eq:route-acc}.

\emph{Random comparator.}
If $\routeset$ is uniform over all subsets of size $M_\budget$, then every item
has inclusion probability $M_\budget/N=b_\budget$. Linearity of expectation gives
\begin{equation*}
  \E_{\routeset}\!\left[
    \sum_{i\in\routeset}\repair_i\right]
  =\sum_{i=1}^{N}\repair_i\Prob_{\routeset}(i\in\routeset)
  =M_\budget\bar\repair.
\end{equation*}
Substitution into \cref*{eq:route-acc} proves \cref*{eq:route-random}. Taking
$\routeset=[N]$ also shows that $\bar\repair$ is the accuracy change produced by
routing every item.

\emph{Oracle comparator.}
Because \cref*{eq:route-acc} is linear in the selected repair values, a size-
$M_\budget$ maximizer selects them in descending order. The realized population
contains $N\prevplus$ values $+1$, $N(1-\prevplus-\prevminus)$ zeros, and
$N\prevminus$ values $-1$. The largest $M_\budget$ values therefore sum to
\begin{equation*}
  \min(M_\budget,N\prevplus)
  -\max\{0,M_\budget-N(1-\prevminus)\}.
\end{equation*}
Dividing by $N$ proves \cref*{eq:route-oracle}. The maximum is first attained at
$b_\budget=\prevplus$, remains constant while zero-valued items are available
($\prevplus\le b_\budget\le1-\prevminus$), and then declines. At
$b_\budget=1$, its gain over greedy is
$\prevplus-\prevminus=\bar\repair$, as required by route-all accuracy.
\end{proof}

The implemented cost is $1+k|\routeset|/N$ solver calls per item: every item
receives one initial greedy solve and each routed item receives $k$ new solves.
Reusing the greedy answer as one of the $k$ votes would define a different cost
and a different experiment.

The corresponding artifact check is reported in \cref{app:numerical-routing}.

\subsection{Numerical Verification}
\label{app:numerical-checks}

These checks ask whether the stored results implement the proved identities.
The algebraic decomposition in \cref*{prop:residual-shift} requires no
data-dependent check.
\Cref{tab:identity-check-summary} gives the conclusions before the detailed
values.

\begin{table}[h]
\centering\small
\caption{Summary of the artifact-level identity checks.}
\label{tab:identity-check-summary}
\begin{tabular}{@{}p{0.25\columnwidth}p{0.64\columnwidth}@{}}
\toprule
Identity & Result \\
\midrule
Residual risk
  & Agreement to floating-point precision in all 42 cells. \\
AUROC
  & Difference equals the exact $1/(2N)$ endpoint term to numerical precision. \\
Routing
  & Oracle curves agree to floating-point precision in both $k=5$ arms. \\
\bottomrule
\end{tabular}
\end{table}

\subsubsection{Residual-risk identity}
\label{app:numerical-residual}

\Cref*{eq:residual-delta} reproduces every recorded residual-risk improvement
from the measured recall gain, prevalence, and realized budget. Agreement is to
floating-point precision over all 42 model--dataset--budget cells. Using the
nominal budget instead of $b_\budget=M_\budget/N$ would create discrepancies up
to $1.6\times10^{-5}$. The six cells at nominal $\budget=0.05$ are:

\begin{center}\small
\setlength{\tabcolsep}{3pt}
\begin{tabular}{lrrrr}
\toprule
Run & $\prev$ & $\Delta\recall$ & predicted & recorded \\
\midrule
Gemma~3/GSM8K   & 0.0607 & $+0.2250$ & $0.0144$ & $0.0144$ \\
Qwen2.5/GSM8K   & 0.0811 & $+0.0935$ & $0.0080$ & $0.0080$ \\
Llama~3.1/GSM8K & 0.1783 & $+0.1073$ & $0.0201$ & $0.0201$ \\
Gemma~3/MATH    & 0.1837 & $+0.0546$ & $0.0106$ & $0.0106$ \\
Qwen2.5/MATH    & 0.2695 & $+0.0135$ & $0.0038$ & $0.0038$ \\
Llama~3.1/MATH  & 0.4944 & $+0.0208$ & $0.0108$ & $0.0108$ \\
\bottomrule
\end{tabular}
\end{center}

\subsubsection{Finite-sample AUROC identity}
\label{app:numerical-auroc}

The exact discrete sum agrees with \cref{eq:auroc-finite} on the aligned
held-out predictions, with ties averaged analytically. To display the endpoint
term, \cref{tab:auroc-check} reports
$\mathrm{RHS}_0:=\prev/2+(1-\prev)\auroc$ without $1/(2N)$; the predicted
difference is therefore $\mathrm{LHS}-\mathrm{RHS}_0=1/(2N)$.

\begin{table}[h]
\centering\small
\caption{Finite-sample check of \cref*{prop:auroc}. LHS
$=N^{-1}\sum_j \recall_{j/N}$; the difference from $\mathrm{RHS}_0$ should be
$1/(2N)$.}
\label{tab:auroc-check}
\begin{tabular}{llrrr}
\toprule
Run & score & $\auroc$ & LHS & $\mathrm{RHS}_0$ \\
\midrule
Gemma~3/GSM8K   & probe & 0.8787 & 0.8561 & 0.8557 \\
Gemma~3/GSM8K   & judge & 0.8484 & 0.8276 & 0.8272 \\
Qwen2.5/GSM8K   & probe & 0.8583 & 0.8296 & 0.8293 \\
Qwen2.5/GSM8K   & judge & 0.8731 & 0.8432 & 0.8428 \\
Llama~3.1/GSM8K & probe & 0.8127 & 0.7574 & 0.7570 \\
Llama~3.1/GSM8K & judge & 0.6969 & 0.6622 & 0.6618 \\
Gemma~3/MATH    & probe & 0.9141 & 0.8381 & 0.8380 \\
Gemma~3/MATH    & judge & 0.8221 & 0.7631 & 0.7630 \\
Qwen2.5/MATH    & probe & 0.8878 & 0.7833 & 0.7832 \\
Qwen2.5/MATH    & judge & 0.8479 & 0.7542 & 0.7541 \\
Llama~3.1/MATH  & probe & 0.8647 & 0.6845 & 0.6844 \\
Llama~3.1/MATH  & judge & 0.6955 & 0.5990 & 0.5989 \\
\bottomrule
\end{tabular}
\end{table}

The displayed difference is at most $4.0\times10^{-4}$. Before rounding, it
equals $1/(2N)\leq 3.9\times10^{-4}$ to numerical precision: the GSM8K rows
($N\approx1300$) lie near $4\times10^{-4}$ and the MATH rows
($N\approx4900$) near $1\times10^{-4}$.

\subsubsection{Routing identities}
\label{app:numerical-routing}

Both $k=5$ arms satisfy the oracle identity to floating-point precision. The
stored \texttt{random} curve is a Monte Carlo permutation average, so it
approximates---rather than exactly equals---the expectation in
\cref*{eq:route-random}.

On Llama~3.1~8B/GSM8K ($N=1308$), five-sample voting repairs $120$ items and
breaks $28$. This is the separately generated $k=5$ voting population, not the
$N=1307$ $k=1$ discrimination population (\cref*{sec:limitations}). Hence
$\bar\repair=(120-28)/1308$, and the exact random-routing
expectation at any realized fraction $b_\budget$ is
$0.8249235+b_\budget(92/1308)$. At nominal $\budget=0.10$, the implemented cost
is approximately $1.50$ solver calls per item; the exact expectation and cost use
the realized $b_\budget$ after rounding. \Cref*{eq:route-oracle} predicts a ceiling
of $0.8249+120/1308=0.9167$, first reached at
$\prevplus=120/1308=0.0917$. The curve remains there while zeros are available,
then falls at route-all to $0.8953$; the ceiling exceeds route-all by
$28/1308=0.0214=\prevminus$.

On Gemma~3~12B/GSM8K ($N=1319$), voting repairs $19$ and breaks $7$, on a much
higher greedy baseline. The predicted ceiling is $0.9416+19/1319=0.9560$ at
$b_\budget=19/1319=0.0144$, against a recorded $0.9560$ on the plateau; it then
falls to route-everything ($0.9507$) at $b_\budget=1$. The ceiling excess is
$0.0053=7/1319=\prevminus$. The oracle identity is exact in both arms across an
eightfold difference in repairable mass. This agreement checks the artifact
implementation; the reason the formula holds is the finite-sample decomposition
above, not a fit to these two curves.

The numerical identity checks are reproduced by
\texttt{scripts/verify\_identities.py}.

\subsection{Scope of the Results}

The exact identities above explain relationships among the paper's reported
quantities; they are not deployment guarantees. In particular,
\cref*{prop:residual-shift} does not bound performance under domain shift, and
\cref{prop:fisher} applies to an unregularised population direction under a
shared-covariance Gaussian model, not exactly to the regularised finite-sample
selector in \cref*{eq:alpha}. Numerical agreement checks the artifact's
implementation of the finite identities but does not replace their proofs.

%% file: appendix/B-protocol.tex
\section{Estimation and Uncertainty Protocol}
\label{app:protocol}

This appendix specifies how every reported estimate and uncertainty interval is
computed.

\subsection{Tie handling at a budget cutoff}
\label{app:ties}

A budget selects an integer $M_\budget$ of items by score (nearest-integer for the
review analysis and floor for routing; \cref*{sec:setup-stats}). When the
cutoff falls strictly inside a group of tied scores, the selected set is not
determined by the score alone, and resolving the ambiguity by sort order makes a
reported number depend on an implementation detail. This can occur for the
black-box baseline because finite-precision $\pyes$ values repeat, although a
near-zero or near-one value is not by itself an exact tie
(\cref{app:pyes-saturation}).

We report the exact expectation under uniform random tie-breaking. Let the cutoff
fall inside a tie group $G$ containing $|G|$ items of which $W_G$ are errors, with
$j$ selection slots remaining after all strictly-lower-scoring items are taken.
The expected number of errors captured from $G$ is $j\,W_G/|G|$, so the
tie-averaged recall is
\begin{equation}
  \label{eq:tie}
  \recall_{\budget} \;=\; \frac{1}{W}\left(W_{<} + \frac{j\,W_G}{|G|}\right),
\end{equation}
with $W_{<}$ the errors among strictly-lower-scoring items. This is the
expectation over the randomisation, computed in closed form rather than by
sampling, and it is applied identically to every detector. The same convention
defines the tie-averaged risk--coverage curves of \cref{app:risk-coverage}.

\subsection{Bootstrap}
\label{app:bootstrap}

Differences between detectors are resampled \emph{paired}: one bootstrap draw
selects a multiset of held-out test items, and every detector is re-scored on that
same multiset, so the resulting interval reflects the item-level variation the two
detectors share rather than treating them as independent samples. Discrimination
and selective-risk metrics use class-stratified resampling, which keeps the number
of minority-class items fixed across draws and prevents an interval from being
dominated by fluctuation in $\prev$; the routing analysis resamples routing items
unstratified and recomputes the ordering within each draw. All resampling uses
$\num{2000}$ draws and fixed seeds recorded in the analysis source and output
caches.

Two properties matter for reading the intervals. First, all fitted objects ---
head set, standardisation, residualiser, and probe ensemble --- and their score
vectors are frozen before resampling. The fixed-budget ranking and cutoff are
reapplied to each resampled multiset. The intervals are therefore
\emph{conditional on the fitted ensemble} and do not include fitting variability.
Second, fitting variability is reported separately rather than pooled: individual
initialisation AUROCs have standard deviations of $0.003$--$0.014$, and the
train-resample spread (which
additionally varies the sampled training set and the selected heads) is reported
alongside it. Combining the two into one $\pm$ would answer neither question.

\subsection{Why the operating point is a budget}
An earlier analysis chose each threshold on train to match the judge's training
precision. That does not equalise test precision after prevalence or score drift,
so the resulting recalls lie at different test operating points. We instead rank
without test labels and require every detector to select the same $M_\budget$
items; this is both operationally defined and comparable across detectors.

\subsection{Proportion intervals}
\label{app:wilson}

Recalls at a fixed budget are proportions of a fixed denominator (the number of
errors in the analysed population), and their marginal intervals use the Wilson
score interval rather than the Wald interval, which is unreliable at the small
error counts of the low-prevalence GSM8K cells (as few as $80$ errors for
Gemma~3/GSM8K). Differences between detectors always use the paired bootstrap of
\cref{app:bootstrap}, never a difference of two marginal intervals.

\subsection{What is not corrected}
\label{app:multiplicity}

The six model--dataset cells constitute a family of comparisons and the reported
intervals are per-cell and nominal. We do not apply a family-wise correction, and
we do not want the reader to infer one has been applied. Similarly, the
hyperparameter ranges of \cref{app:sweep} are minima and maxima over a grid
evaluated on the test split; they are descriptive extrema, not confidence
intervals, and they do not license selecting a configuration post hoc. The MATH
difficulty strata of \cref{app:levels} carry no interval at all.

\subsection{Why no calibration metric is reported}
\label{app:calibration}

Every quantity in this paper is a functional of the ordering induced by the score
(\cref*{sec:setting-levers}), so any strictly increasing reparametrisation leaves
all results unchanged. The probe's outputs are consequently \emph{scores}, not
calibrated probabilities, and reporting expected calibration error or a Brier
score would attach probability semantics the construction does not support: the
MLP oversamples the minority class and the logistic regression uses balanced class
weights, both of which shift the output scale without a separate train-only
calibration stage. A deployment that needs a calibrated probability rather than a
ranking would have to add that stage and validate it separately.

%% file: appendix/C-pyes-audit.tex
\section{Audit of the \texorpdfstring{$\pyes$}{p(YES)} Baseline}
\label{app:pyes}

The comparison in this paper is only as sound as its baseline, so the construction
of \cref*{eq:pyes} is audited rather than assumed. Three things could make the
baseline unfairly weak: a bad verdict-token rule, a distribution the deployer
could not actually observe, and saturation that a careless cutoff would resolve
arbitrarily. We measure all three.

\subsection{The verdict-token rule}
\label{app:pyes-rule}

$V_{\yes}$ and $V_{\no}$ are built by matching the verdict-position vocabulary
against surface forms of the two verdicts. On a subword vocabulary this is
error-prone in an uncontrolled direction: on Gemma~3's vocabulary, $765$ distinct
tokens match a \no\ prefix under a naive rule, including \texttt{"~not"} and
\texttt{"~November"}. We therefore compare prefix matching against exact matching
of the verdict tokens. The two agree to within $1.1\times10^{-3}$ AUROC in the
worst of the six cells (Llama~3.1/MATH, $0.6956$ prefix versus $0.6946$ exact) and
to within $10^{-5}$ in the best, so the choice of rule does not carry any
conclusion in the paper. The rule is pinned by a regression test and the per-model
matched-token counts are recorded in the artifact manifest.

The recomputed $\pyes$ is additionally cross-checked against the value stored at
generation time for every record; across all runs reported here the two agree
exactly (maximum absolute difference $0$). That check is not decorative --- it
caught a stale verdict-distribution dump that survived every configuration- and
version-based cache check, because those fingerprint the registry entry and the
metric formula but not the dump's contents.

\subsection{What a real API returns}
\label{app:pyes-topk}

A commercial API exposes only the top-$K$ log-probabilities, typically $K=20$,
not the full next-token distribution, so a baseline computed from the full
distribution is stronger than one a black-box deployer could build. We recompute
\cref*{eq:pyes} under top-$20$ and top-$40$ caps, renormalising over the verdict
tokens that survive the cap and assigning the neutral value $0.5$ where neither
appears.

The baseline is essentially unaffected: the largest AUROC difference between the
capped and uncapped forms across the six cells is $9.2\times10^{-5}$
(Llama~3.1/GSM8K), and the fallback to $0.5$ fires on at most $7$ of $4883$
records (Gemma~3/MATH) and on zero records in four of the six cells. The verdict
tokens are, in practice, the top of the distribution at the verdict position. We
report the uncapped form in the main text, which is the marginally stronger
baseline and therefore the conservative choice for our claim.

\subsection{Saturation and exact tie sizes}
\label{app:pyes-saturation}

$\pyes$ is concentrated near the endpoints. Counting records with $\pyes\le0.02$ or
$\pyes\ge0.98$: $1270$ of $1319$ (Gemma~3/GSM8K), $1260$ of $1319$
(Qwen2.5/GSM8K), $977$ of $1307$ (Llama~3.1/GSM8K), $4654$ of $4883$
(Gemma~3/MATH), $4408$ of $4949$ (Qwen2.5/MATH), and $3039$ of $4466$
(Llama~3.1/MATH). Thus $68\%$--$96\%$ lie near an endpoint, but this saturation
band is not an exact tie group.

The exact score audit gives 144, 137, and 688 unique values on the three GSM8K
cells, and 418, 668, and 1686 on MATH (Gemma~3, Qwen2.5, Llama~3.1). The largest
exact tie groups have 64, 58, 17, 173, 162, and 74 items respectively. At the 5\%
cutoff, the corresponding tie groups contain only 2, 2, 1, 1, 1, and 1 items; over
all reported review budgets the largest cutoff tie contains 56 items. Exact
tie-averaging therefore remains the reproducible convention, but its effect is
localized rather than the thousand-item ambiguity previously implied by the
saturation counts.

\subsection{Why the sampled verdict is not the baseline}
\label{app:pyes-verdict}

The binary verdict $v\in\{\yes,\no\}$ is a thresholded read of \cref*{eq:pyes}, and
its ROC area is algebraically balanced accuracy,
$\tfrac{1}{2}(\mathrm{TPR}+\mathrm{TNR})$. This is a valid AUROC for a two-level
ranking, but it discards all within-verdict ordering. Using the operational
$\pyes$ values reported in \cref*{tab:main}, the gap between the two forms of the
same signal is large: $0.703$ versus $0.848$ (Gemma~3/GSM8K), $0.582$ versus $0.873$
(Qwen2.5/GSM8K), $0.545$ versus $0.697$ (Llama~3.1/GSM8K), and $0.602$, $0.614$,
$0.522$ versus $0.822$, $0.848$, $0.696$ on the three MATH runs. Comparing a probe
against the verdict rather than against $\pyes$ would therefore credit the probe
with up to $0.29$ AUROC that is attributable to reading a distribution instead
of a token. We therefore retain the verdict only as this diagnostic and use
$\pyes$ for every operational comparison.

%% file: appendix/D-populations.tex
\section{Analysed Populations and Exclusions}
\label{app:populations}

\input{tables/table4_populations}

\Cref{tab:populations} gives the full exclusion accounting behind
\cref*{sec:setup-data}. Three points about it are worth stating explicitly.

\paragraph*{The exclusion changes the target population}
A budget-truncated solution stops before writing its final answer, so it parses to
no answer and would be scored wrong. Retaining these records would inflate the
observed error prevalence --- from $0.312$ to $0.371$ in the Llama~3.1/MATH
train split, for instance --- and add a potentially easy-to-detect population of
texts that stop mid-sentence. Excluding them lowers prevalence in every affected
cell (the $\prev$ versus $\prev^{+}$ columns), but the direction of its effect on
the probe--judge difference is not known.

\paragraph*{Llama's MATH exclusions are non-termination, not long solutions}
Llama~3.1~8B loses $8.6\%$ of its MATH train split and $10.7\%$ of its test split,
by far the largest exclusion in the table, and this shifts its analysed prevalence
from $0.312$ (train) to $0.494$ (test). A budget-calibration run finds that $37\%$
of its truncated problems still exceed $12{,}288$ tokens --- three times the
budget actually used --- so these are generations that do not terminate rather
than ordinary long solutions. Excluding them defines performance conditional on
termination; a system-level evaluation could instead count non-termination as
failure. The train/test prevalence shift
nonetheless means Llama's MATH probe is fitted at a materially different error rate
than it is evaluated at, which is a caveat on that cell specifically.

\paragraph*{Labels are recomputed, not trusted}
Each record carries an advisory \texttt{labelable} field written at generation
time, and the authoritative value is recomputed at analysis time. Files written
before this was corrected recorded the field as ``a gold answer exists'', ignoring
truncation, and therefore disagreed with the analysis on exactly the truncated
records. The loader audits the two and prints any divergence rather than allowing
a hand inspection of the raw records to imply a different population than the
figures were built from.

\subsection{Answer equivalence}
\label{app:equivalence}

The relation $\equiv$ of \cref*{eq:label} is per-dataset and is recomputed from
each raw record at analysis time.

\emph{GSM8K} uses exact numeric equivalence after parsing, so \texttt{64},
\texttt{64.0} and \texttt{64.00} are one answer.

\emph{MATH} normalises \LaTeX{} and then compares as strings, additionally
accepting decimal equality when \emph{both} sides are plain numeric literals.
Thus \verb|\dfrac{1}{2}|, \verb|\frac12| and \verb|\frac{1}{2}| are one answer,
while \verb|\frac{1}{2}| and \verb|0.5| are \emph{not}. This strict rule may
mislabel equivalent forms and is not accompanied by a symbolic-equivalence
sensitivity. The solve prompt requests the final answer inside
\verb|\boxed{}|, and reference answers are read from the last \verb|\boxed{}| of
the reference solution; a reference that boxes nothing is unlabelable and
excluded, whereas a \emph{model} response that boxes nothing counts as wrong,
because the reference answer is still known.

%% file: tables/table4_populations.tex
\begin{table}[t]
\centering
\caption{Analysed population and exclusions, per model--dataset cell and split. ``trunc'' is
generations cut off by the token budget before writing a final answer;
``no gold'' is reference answers that did not parse. $\prev$ is the error
prevalence of the analysed population and $\prev^{+}$ what it would be if
truncations were retained and scored wrong. Excluding them lowers observed
prevalence in every affected cell.}
\label{tab:populations}
\scriptsize
\setlength{\tabcolsep}{2pt}
\begin{tabular}{llrrrrrr}
\toprule
Run & split & gen. & trunc & no gold & analysed & $\prev$ & $\prev^{+}$ \\
\midrule
Gemma~3 12B / GSM8K & train & 7473 & 7 & 0 & 7466 & 0.057 & 0.057 \\
 & test & 1319 & 0 & 0 & 1319 & 0.061 & 0.061 \\
Qwen2.5 7B / GSM8K & train & 7473 & 0 & 0 & 7473 & 0.050 & 0.050 \\
 & test & 1319 & 0 & 0 & 1319 & 0.081 & 0.081 \\
Llama~3.1 8B / GSM8K & train & 7473 & 30 & 0 & 7443 & 0.107 & 0.110 \\
 & test & 1319 & 12 & 0 & 1307 & 0.178 & 0.185 \\
\midrule
Gemma~3 12B / MATH & train & 7500 & 197 & 2 & 7301 & 0.177 & 0.198 \\
 & test & 5000 & 117 & 0 & 4883 & 0.184 & 0.203 \\
Qwen2.5 7B / MATH & train & 7500 & 83 & 2 & 7415 & 0.234 & 0.242 \\
 & test & 5000 & 51 & 0 & 4949 & 0.270 & 0.277 \\
Llama~3.1 8B / MATH & train & 7500 & 645 & 2 & 6853 & 0.312 & 0.371 \\
 & test & 5000 & 534 & 0 & 4466 & 0.494 & 0.548 \\
\midrule
\textbf{total} & & 63876 & 1676 & 6 & 62194 & & \\
\bottomrule
\end{tabular}
\end{table}


%% file: appendix/E-ablations.tex
\section{Robustness and Bounds}
\label{app:ablations}

These analyses support the main claims rather than carrying them. Each is labelled
with what it can and cannot establish.

\input{tables/table5_transfer_full}

\begin{figure*}[t]
  \centering
  \includegraphics[width=\textwidth]{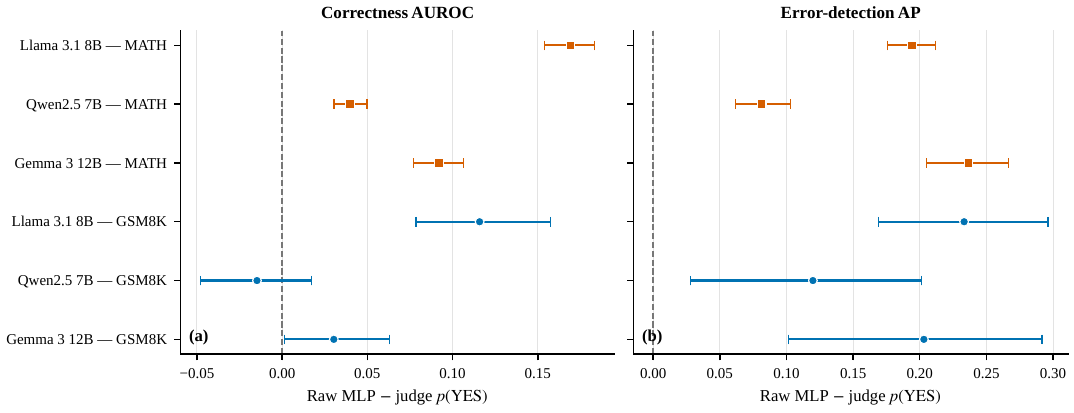}
  \caption{Probe minus judge $\pyes$ on held-out correctness AUROC (left) and
  error-class average precision (right), with paired 95\% class-stratified
  bootstrap intervals over test items. AP treats errors as the positive class and
  is therefore sensitive to the observed prevalence. Intervals are conditional on
  the fitted five-seed ensemble. These are the values tabulated in
  \cref*{tab:main}.}
  \label{fig:discrimination}
\end{figure*}

\begin{figure*}[t]
  \centering
  \includegraphics[width=\textwidth]{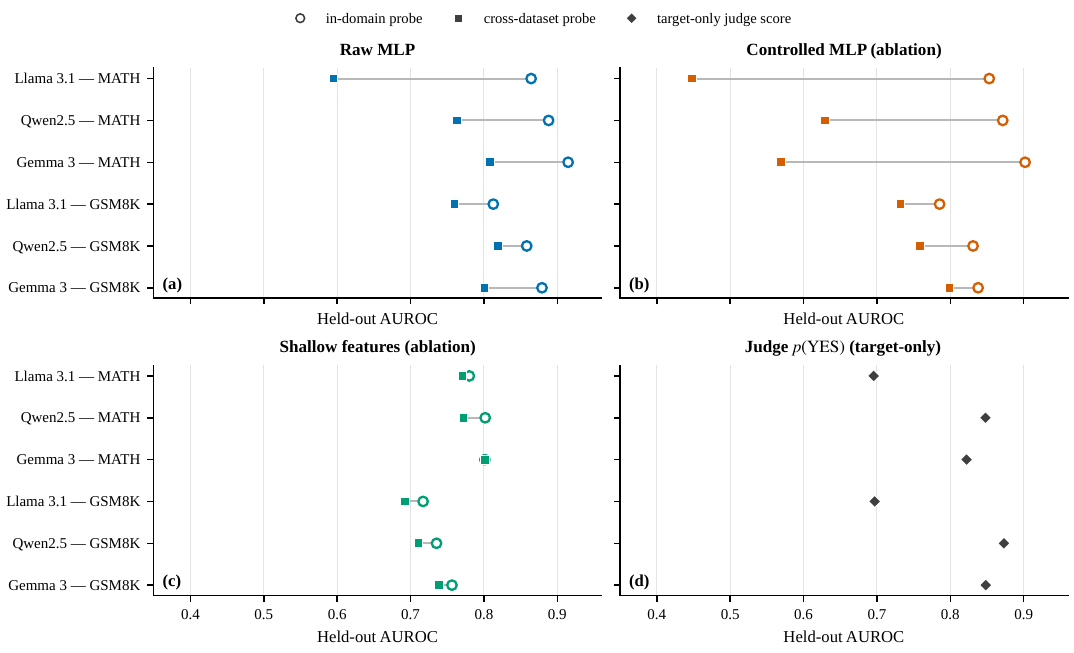}
  \caption{Held-out target-split AUROC when each score is fitted in domain (open
  circle) or on the other mathematics dataset (filled square). Diamonds mark the
  target-domain judge $\pyes$, which needs no probe fitting. Paired intervals for
  every difference shown here are in \cref{tab:transfer-full}.}
  \label{fig:transfer}
\end{figure*}

\subsection{Full transfer matrix}
\label{app:transfer-full}

\Cref{tab:transfer-full} gives every fit--target direction for all three
detectors with paired intervals, of which \cref*{tab:transfer} shows the primary
probe, and \cref{fig:transfer} plots it. Two patterns carry the argument of
\cref*{sec:trust-boundary}. The
controlled probe's loss exceeds the raw probe's on the same items in every
GSM8K$\to$MATH direction, a pattern consistent with
\cref*{prop:residual-shift}. And the shallow detector loses least of the three in all six
directions, with three of its six loss intervals including zero.

\subsection{Full risk--coverage profiles}
\label{app:risk-coverage}

\begin{figure*}[t]
  \centering
  \includegraphics[width=\textwidth]{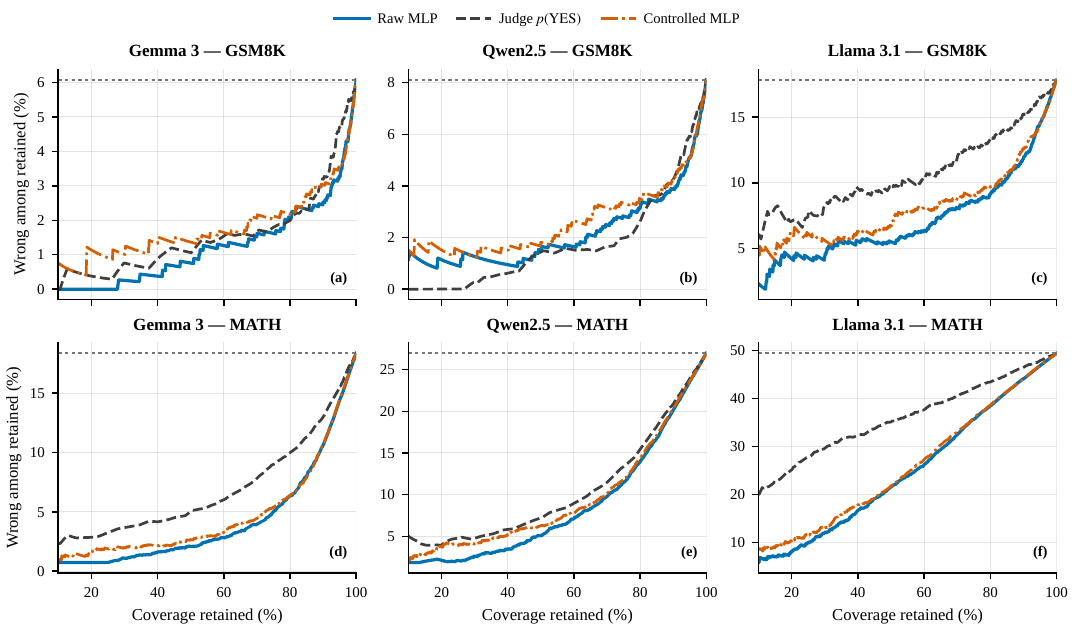}
  \caption{Wrong-answer rate among retained answers as realised coverage
  $c_\budget=1-b_\budget$
  increases from $10\%$ to $100\%$; lower is better. The dashed line is the
  no-filtering prevalence, which every curve meets at $c=1$ by construction. Ties
  use their exact expected risk under uniform random tie-breaking
  (\cref{app:ties}); exact cutoff-tie sizes are audited in
  \cref{app:pyes-saturation}.}
  \label{fig:risk-coverage}
\end{figure*}

\Cref{fig:risk-coverage} gives the finite-sample selective-prediction view,
$\srisk(c_\budget)=\resid_\budget$, over the full coverage range. It is the same measurement
as \cref*{fig:budget} by \cref*{prop:residual} and is included because it is the
conventional presentation and because it shows the whole curve rather than the
deployable region. Note that reading a summary such as $\aurc$ off this figure
integrates uniformly over coverage --- the same objection \cref*{prop:auroc} raises
against AUROC.

\subsection{Evidence consistent with a nonlinear read-out advantage}
\label{app:sweep}

\begin{figure*}[t]
  \centering
  \includegraphics[width=\textwidth]{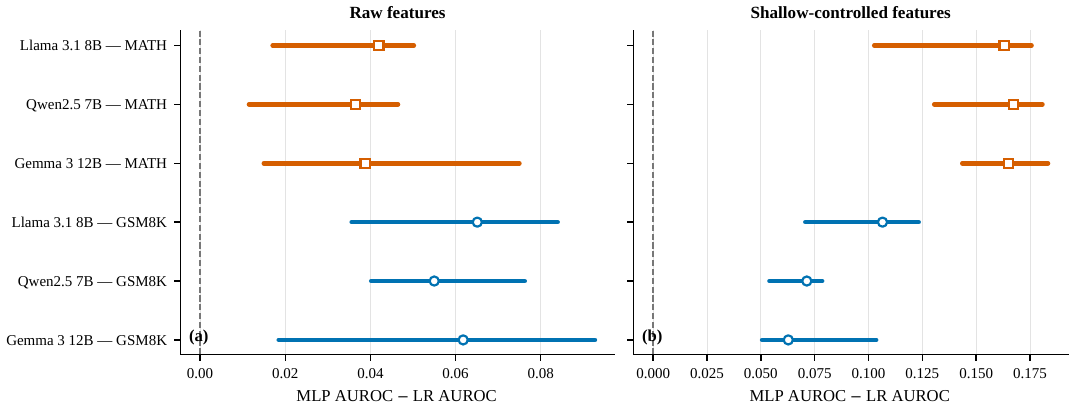}
  \caption{Range of held-out MLP-minus-LR AUROC as the number of retained heads,
  regularisation, and hidden width vary; open circles mark the frozen default.
  These are descriptive test-set extrema, \emph{not} confidence intervals, and
  they do not license selecting a configuration post hoc or attributing the gap
  to nonlinearity alone.}
  \label{fig:sweep}
\end{figure*}

The three arbitrary defaults --- $n_{\mathrm h}=32$ heads, $C=0.5$, one hidden
layer of $128$ units --- were varied one axis at a time
($n_{\mathrm h}\in[8,128]$, $C\in[0.05,10]$, hidden
sizes from $(32)$ to $(128,128)$) across all six model--dataset cells, giving $84$ cells. The MLP
beats logistic regression in all $84$. Granting each probe family its most
favourable cell anywhere in the grid --- which hands LR its best configuration and
is therefore the comparison least flattering to our claim --- still leaves the MLP
ahead by $+0.054$ to $+0.156$ AUROC.

Two details bear on the design choices rather than on the headline. Shrinking the
hidden layer from $128$ to $32$ units costs under $0.009$ AUROC, so the observed
gap is not strongly sensitive to hidden width in this grid. The default
$n_{\mathrm h}=32$ \emph{understates} the linear
probe: LR climbs monotonically as more heads are added while the MLP is flat in
$n_{\mathrm h}$, so a larger $n_{\mathrm h}$ would narrow the gap on the linear
side rather than widen the nonlinear one. The comparison is consistent with a nonlinear read-out advantage,
but MLP and LR also differ in ensembling, minority balancing, optimisation, and
regularisation, so it does not isolate nonlinearity causally. The monotone
behaviour of LR in $n_{\mathrm h}$ is also why we make no sparsity claim in
\cref*{sec:method-heads}.

\subsection{Surface-form perturbation}
\label{app:symbolic}

\begin{figure}[t]
  \centering
  \includegraphics[width=\columnwidth]{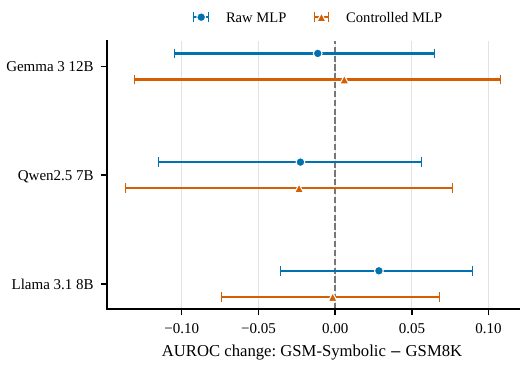}
  \caption{AUROC difference between GSM-Symbolic and the full GSM8K test split.
  Whiskers independently resample template clusters on the perturbed side and
  rows on the full-test side; this is not a paired comparison. Intervals generally
  include zero, so the figure bounds only a large population difference and does
  not establish perturbation invariance.}
  \label{fig:symbolic}
\end{figure}

GSM8K templates re-instantiated with fresh names and numbers give new benchmark
instances, but their absence from pretraining cannot be established. Probes are fitted on the real
GSM8K train split only and scored on the perturbed set. Primary-probe AUROC moves
by $-0.023$ (Qwen2.5), $+0.029$ (Llama~3.1) and $-0.011$ (Gemma~3) against the
full GSM8K test split, and solver accuracy moves by $-4.6$, $-0.7$ and $-1.8$
points paired over templates.

\paragraph*{Read this as a bound, not an exclusion.}
Every one of those intervals covers zero and is roughly $\pm0.1$ AUROC wide,
because the perturbed side is $100$ templates however many instances they
generate. The perturbation rules out a \emph{large} recognition effect and cannot
rule out a small one; a few hundredths of AUROC would pass unseen. Tightening the
bound requires more templates, not more instances per template. GSM-Symbolic's
templates also derive from GSM8K, so absence from pretraining is not established
even for the perturbed set.

\paragraph*{What is resolvable.}
The probe-minus-judge margin \emph{on the perturbed set} is paired row by row and
is measured much more sharply than either detector's absolute shift. For
Llama~3.1 it is $+0.147$ $[+0.085,+0.216]$, excluding zero: the probe's advantage
survives re-instantiation wherever it held on the original GSM8K.

\paragraph*{An observation we decline to interpret.}
The largest movement anywhere in the perturbation table belongs to Qwen2.5's
judge, whose $\pyes$ AUROC falls from $0.873$ to $0.783$ ($-0.090$). Its interval
is $[-0.208,+0.016]$ and covers zero. The paired difference-in-differences against
the probe, which cancels most of the resampling noise, is $+0.067$
$[-0.020,+0.141]$ and also covers zero; only against shallow text features does it
exclude zero ($+0.119$ $[+0.032,+0.210]$), and that is partly because shallow
features \emph{gain} on the perturbed set while the judge loses. Llama~3.1
($0.697/0.696/0.694$ across GSM8K, MATH and perturbed GSM8K) and Gemma~3
($0.848/0.822/0.865$) are stable across all three. The supportable statement is
that Qwen2.5's judge behaves differently from every other detector under
re-instantiation; benchmark familiarity is a hypothesis this experiment does not
have the resolution to test.

\subsection{Difficulty strata}
\label{app:levels}

Stratifying the MATH test splits by the dataset's own difficulty level gives $15$
strata ($3$ models $\times$ $5$ levels, $n=419$ to $1298$) spanning error
prevalences from $3.7\%$ to $74.8\%$. The probe-minus-judge margin is positive in
all $15$, ranging from $+0.021$ (Qwen2.5, Level~4) to $+0.188$ (Llama~3.1,
Level~1), and it remains positive at Level~5 for every model ($+0.045$, $+0.154$,
$+0.091$). Primary-probe AUROC itself is near flat across the range, from $0.810$
to $0.936$, despite the twentyfold change in prevalence.

This is descriptive: there are no level-wise intervals, the strata are not
independent of the model's own competence, and difficulty is the dataset's
annotation rather than a measured quantity. What it supports is a negative claim ---
the probe's advantage is not confined to easy problems, and does not vanish where
the model fails most --- which is the relevant one for the contamination question
of \cref*{sec:limitations}.

%% file: tables/table5_transfer_full.tex
\begin{table*}[t]
\centering
\caption{Full transfer matrix under the joint protocol change. Columns as in \cref*{tab:transfer};
$\dagger$ marks a paired 95\% interval excluding zero. The controlled probe loses
more than the raw probe on the same items in every GSM8K$\to$MATH direction,
a pattern consistent with \cref*{prop:residual-shift} but not a causal test; the shallow detector loses
least of the three everywhere.}
\label{tab:transfer-full}
\footnotesize
\setlength{\tabcolsep}{4pt}
\begin{tabular}{llrlrrrr}
\toprule
Model & Target & judge & detector & in & $\to$ & $\Delta$ vs.\ in-dom.\ [95\% CI] & $\Delta$ vs.\ judge [95\% CI] \\
\midrule
Gemma~3 12B & GSM8K & 0.848 & raw MLP & 0.879 & 0.808 & -0.071~[-0.109, -0.034]$^{\dagger}$ & -0.040~[-0.077, -0.002]$^{\dagger}$ \\
 &  &  & controlled & 0.844 & 0.802 & -0.041~[-0.097, +0.017]\phantom{$^{\dagger}$} & -0.046~[-0.093, -0.002]$^{\dagger}$ \\
 &  &  & shallow & 0.756 & 0.738 & -0.018~[-0.048, +0.013]\phantom{$^{\dagger}$} & -0.109~[-0.168, -0.054]$^{\dagger}$ \\
\addlinespace[1.5pt]
 & MATH & 0.822 & raw MLP & 0.914 & 0.808 & -0.106~[-0.122, -0.091]$^{\dagger}$ & -0.014~[-0.027, -0.001]$^{\dagger}$ \\
 &  &  & controlled & 0.902 & 0.567 & -0.335~[-0.360, -0.310]$^{\dagger}$ & -0.255~[-0.280, -0.230]$^{\dagger}$ \\
 &  &  & shallow & 0.801 & 0.801 & -0.000~[-0.005, +0.005]\phantom{$^{\dagger}$} & -0.021~[-0.035, -0.007]$^{\dagger}$ \\
\addlinespace[1.5pt]
\midrule
Qwen2.5 7B & GSM8K & 0.873 & raw MLP & 0.858 & 0.819 & -0.040~[-0.080, -0.001]$^{\dagger}$ & -0.054~[-0.092, -0.020]$^{\dagger}$ \\
 &  &  & controlled & 0.830 & 0.774 & -0.057~[-0.109, -0.005]$^{\dagger}$ & -0.100~[-0.148, -0.052]$^{\dagger}$ \\
 &  &  & shallow & 0.735 & 0.711 & -0.025~[-0.054, +0.002]\phantom{$^{\dagger}$} & -0.162~[-0.208, -0.120]$^{\dagger}$ \\
\addlinespace[1.5pt]
 & MATH & 0.848 & raw MLP & 0.888 & 0.763 & -0.125~[-0.139, -0.110]$^{\dagger}$ & -0.085~[-0.099, -0.070]$^{\dagger}$ \\
 &  &  & controlled & 0.873 & 0.624 & -0.249~[-0.268, -0.230]$^{\dagger}$ & -0.224~[-0.244, -0.203]$^{\dagger}$ \\
 &  &  & shallow & 0.802 & 0.772 & -0.030~[-0.038, -0.022]$^{\dagger}$ & -0.076~[-0.090, -0.062]$^{\dagger}$ \\
\addlinespace[1.5pt]
\midrule
Llama~3.1 8B & GSM8K & 0.697 & raw MLP & 0.813 & 0.757 & -0.055~[-0.096, -0.016]$^{\dagger}$ & +0.060~[+0.023, +0.099]$^{\dagger}$ \\
 &  &  & controlled & 0.785 & 0.723 & -0.062~[-0.107, -0.016]$^{\dagger}$ & +0.027~[-0.013, +0.066]\phantom{$^{\dagger}$} \\
 &  &  & shallow & 0.717 & 0.692 & -0.025~[-0.045, -0.005]$^{\dagger}$ & -0.005~[-0.053, +0.045]\phantom{$^{\dagger}$} \\
\addlinespace[1.5pt]
 & MATH & 0.696 & raw MLP & 0.865 & 0.595 & -0.270~[-0.289, -0.251]$^{\dagger}$ & -0.100~[-0.117, -0.084]$^{\dagger}$ \\
 &  &  & controlled & 0.853 & 0.448 & -0.405~[-0.426, -0.385]$^{\dagger}$ & -0.247~[-0.266, -0.228]$^{\dagger}$ \\
 &  &  & shallow & 0.780 & 0.771 & -0.009~[-0.015, -0.003]$^{\dagger}$ & +0.075~[+0.057, +0.094]$^{\dagger}$ \\
\addlinespace[1.5pt]
\bottomrule
\end{tabular}
\end{table*}

%% file: appendix/G-reproducibility.tex
\section{Reproducibility and Disclosure}
\label{app:repro}

\subsection{What is released}

The public software and data artifact is available at
\url{https://github.com/MakGulati/auditing-the-judge-artifact}.
The fixed camera-ready release is linked from the start of this supplement.
It contains the generation, judging, activation-extraction, probe-fitting,
routing, transfer, and plotting source; pinned dependency manifests and tests;
portable aligned held-out prediction arrays; cached statistics and rendered
figures; and provenance metadata. It deliberately excludes the manuscript
source, raw generation records, activation dumps, and model weights. The omitted
experimental outputs are regenerable by the supplied scripts from the named
upstream datasets and checkpoints, but are not redistributed because of size and
upstream terms. The artifact therefore supports CPU reproduction of the reported
numerical analyses and figures, as well as end-to-end regeneration when model
access and suitable GPU resources are available.

\subsection{Provenance binding}

Three checks bind an analysis to the generation run it claims to describe, and
each was added after a failure it now prevents.

\emph{Prompt reproduction.} Activation extraction re-templates the judge prompt
from scratch, so a different checkpoint, tokenizer, or assistant prefill would
read activations at a different token position than the one that emitted the
recorded verdict, with no visible symptom downstream. An extractor refuses to run
when it cannot reproduce the prompt.

\emph{Row-level fingerprints.} Every dump row carries a content fingerprint of the
record it was extracted from, so a dump cannot be resumed or analysed against a
raw file it did not come from even when the indices line up.

\emph{Geometry.} Each dump records the model id and attention geometry, and the
probe refuses a head dimension that disagrees with it. A wrong head dimension
previously reshaped silently, fragmenting real attention heads into halves.

Dumps that predate these checks are refused by the loader rather than plotted, and
this --- not a selection decision --- is why two earlier checkpoints are absent from
\cref*{tab:main} (\cref*{sec:setup-models}).

\subsection{Two independent pipelines agree}

The results in this paper were computed by a figure package that fixes the primary
method in advance. A separate, earlier analysis pipeline computes overlapping
quantities by independent code paths, and where the two overlap they agree: the
$5\%$-budget recalls for Gemma~3/GSM8K are $0.4875$ / $0.2625$ / $+0.225$ in both;
the routing gain over random at $\budget=0.10$ for the controlled MLP is $+2.05$
points in both, with bootstrap intervals $[+1.20,+2.96]$ and $[+1.25,+3.00]$
differing only by resampling noise; and per-run cached AUROC agrees to
$6\times10^{-5}$ (raw) and $2\times10^{-3}$ (controlled). We report this because a
cross-implementation check is evidence that the numbers are properties of the data
rather than of one script.

\subsection{Known sources of run-to-run variation}

MLP fits vary by about $0.002$ AUROC between otherwise identical runs, traceable
to BLAS threading; this is below the reported initialisation spread of
$0.003$--$0.014$ but above zero, so numbers recomputed by a different script may
differ in the third decimal. Greedy decoding is only about $47\%$
byte-reproducible across runs at all, which bounds what any single-run absolute
accuracy claim can mean and is why the $k=1$ and $k=5$ populations are reported
separately (\cref*{sec:limitations}). The rebuilt probe scores use a newer
scikit-learn than the earlier caches; the per-run metadata records the full
analysis stack for both. The reported generation runs used a single NVIDIA
RTX~5000 Ada Generation GPU with 32\,GiB memory; the pinned vLLM environment was
verified with driver 555.

\subsection{Probe optimisation settings}

The head covariance uses ridge $\lambda=10^{-2}$. Logistic regression uses
$\ell_2$ regularisation with $C=0.5$ and balanced class weights. The MLP uses an
$\ell_2$ penalty of $10^{-4}$, learning rate $10^{-3}$, at most $300$
iterations, and a 15\% internal early-stopping split; its minority class is
oversampled with replacement. The five ensemble initialisations use seeds
$0$--$4$.